\newif\ifarXive
\arXivetrue

\ifarXive
    \documentclass[acmtog]{acmart}
\else 
    \documentclass[acmtog,anonymous,review]{acmart}
\fi

\acmSubmissionID{296}

\usepackage{booktabs} % For formal tables
\usepackage{amsmath}
\usepackage{physics}
\usepackage{url}
\usepackage{color}
\usepackage{tabularx} % in the preamble

\usepackage[ruled]{algorithm2e} % For algorithms

\SetAlFnt{\small}
\SetAlCapFnt{\small}
\SetAlCapNameFnt{\small}
\SetAlCapHSkip{0pt}

\acmJournal{TOG}
\setcopyright{acmlicensed}\acmJournal{TOG}
\acmYear{2021}\acmVolume{40}\acmNumber{6}\acmArticle{215}\acmMonth{12} \acmDOI{10.1145/3478513.3480537}
\renewcommand\footnotetextcopyrightpermission[1]{}

\begin{document}
	% Title portion
	\title{Barbershop: GAN-based Image Compositing using Segmentation Masks}
	
	% DO NOT ENTER AUTHOR INFORMATION FOR ANONYMOUS TECHNICAL PAPER SUBMISSIONS TO
	%SIGGRAPH 2019!
	
	\author{Peihao Zhu}
	% \orcid{0000-0002-7122-1551}
	% \author{Rameen Abdal}
	% \email{rameen.abdal@kaust.edu.sa}
	\affiliation{%
		\institution{KAUST}
		\country{Saudi Arabia}
	}
	\email{peihao.zhu@kaust.edu.sa}
	
	\author{Rameen Abdal}
	\affiliation{%
		\institution{KAUST}
		\country{Saudi Arabia}
	}
	\email{rameen.abdal@kaust.edu.sa}
	
	\author{John Femiani}
	\affiliation{%
		\institution{Miami University}
		\streetaddress{510 E. High St}
		\city{Oxford}
		\state{OH}
		\postcode{45056}
		\country{USA}
	}
	\email{femianjc@miamioh.edu}
	
	\author{Peter Wonka}
	\affiliation{%
		\institution{KAUST}
		\country{Saudi Arabia}
	}
	\email{pwonka@gmail.com}
	
	% \author{Gang Zhou}
	% \orcid{1234-5678-9012-3456}
	% \affiliation{%
	%  \institution{College of William and Mary}
	%  \streetaddress{104 Jamestown Rd}
	%  \city{Williamsburg}
	%  \state{VA}
	%  \postcode{23185}
	%  \country{USA}}
	% \email{gang_zhou@wm.edu}
	% \author{Valerie B\'eranger}
	% \affiliation{%
	%  \institution{Inria Paris-Rocquencourt}
	%  \city{Rocquencourt}
	%  \country{France}
	% }
	% \email{beranger@inria.fr}
	% \author{Aparna Patel}
	% \affiliation{%
	% \institution{Rajiv Gandhi University}
	% \streetaddress{Rono-Hills}
	% \city{Doimukh}
	% \state{Arunachal Pradesh}
	% \country{India}}
	% \email{aprna_patel@rguhs.ac.in}
	% \author{Huifen Chan}
	% \affiliation{%
	%  \institution{Tsinghua University}
	%  \streetaddress{30 Shuangqing Rd}
	%  \city{Haidian Qu}
	%  \state{Beijing Shi}
	%  \country{China}
	% }
	% \email{chan0345@tsinghua.edu.cn}
	% \author{Ting Yan}
	% \affiliation{%
	%  \institution{Eaton Innovation Center}
	%  \city{Prague}
	%  \country{Czech Republic}}
	% \email{yanting02@gmail.com}
	% \author{Tian He}
	% \affiliation{%
	%  \institution{University of Virginia}
	%  \department{School of Engineering}
	%  \city{Charlottesville}
	%  \state{VA}
	%  \postcode{22903}
	%  \country{USA}
	% }
	% \affiliation{%
	%  \institution{University of Minnesota}
	%  \country{USA}}
	% \email{tinghe@uva.edu}
	% \author{Chengdu Huang}
	% \author{John A. Stankovic}
	% \author{Tarek F. Abdelzaher}
	% \affiliation{%
	%  \institution{University of Virginia}
	%  \department{School of Engineering}
	%  \city{Charlottesville}
	%  \state{VA}
	%  \postcode{22903}
	%  \country{USA}
	% }
	% \renewcommand\shortauthors{Zhou, G. et al}

\definecolor{dartmouthgreen}{rgb}{0.05, 0.5, 0.06}
\definecolor{darkblue}{rgb}{0.0, 0., 0.6}

\ifarXive
    \def\reviii{}
    \def\revii{}
    \def\revii{}
    \def\revision{}
    \def\UrlOfRepo{\url{https://zpdesu.github.io/Barbershop}\xspace}
\else 
    \def\reviii{\color{dartmouthgreen}}
    \def\revii{\color{darkblue}}
    \def\revision{\color{darkblue}\xspace}
    \def\UrlOfRepo{\url{https://authorname.github.io/Barbershop}\xspace}
\fi

% The segmentation mask
\def\Mask{\ensuremath{\mathbf{M}}\xspace}
\def\MaskK{\ensuremath{\mathbf{M}_k}\xspace}

\def\Segment{\ensuremath{\textsc{Segment}}\xspace}
\def\Generator{\ensuremath{G}\xspace}

% The images z(x, y) for FG, BG, and blended images
\def\Image{\ensuremath{\mathbf{I}}\xspace}
\def\BlendedImage{\ensuremath{\Image^\text{blend}}\xspace}
\def\RefImageK{\ensuremath{\Image_k}\xspace}
\def\AlignedRefImageK{\ensuremath{\hat{\Image}_k}\xspace}

% The latent codes (combined, spatial, and style)
\def\Code{\ensuremath{\mathbf{C}}\xspace}
\def\StructureTensor{\ensuremath{\mathbf{F}}\xspace}
\def\AppearanceCode{\ensuremath{\mathbf{S}}\xspace}

\def\CodeK{\ensuremath{\Code_k}\xspace}
\def\StructureTensorK{\ensuremath{\StructureTensor_k}\xspace}
\def\AppearanceCodeK{\ensuremath{\AppearanceCode_k}\xspace}

\def\StructureTensorKInit{\ensuremath{\StructureTensorK^\text{init}}\xspace}

\def\CodeKRec{\ensuremath{\CodeK^\text{rec}}\xspace}
\def\StructureTensorKRec{\ensuremath{\StructureTensorK^\text{rec}}\xspace}

\def\AppearanceCodeKRec{\ensuremath{\AppearanceCodeK}\xspace}

\def\CodeKAlign{\ensuremath{\CodeK^\text{align}}\xspace}
\def\StructureTensorKAlign{\ensuremath{\StructureTensorK^\text{align}}\xspace}

\def\AppearanceCodeKAlign{\ensuremath{\AppearanceCodeK}\xspace}

\def\CodeBlend{\ensuremath{\Code^\text{blend}}\xspace}
\def\StructureTensorBlend{\ensuremath{\StructureTensor^\text{blend}}\xspace}
\def\AppearanceCodeBlend{\ensuremath{\AppearanceCode^\text{blend}}\xspace}

\def\StyleGANTwo{StyleGAN2\xspace}
\def\StyleGAN{StyleGAN\xspace}

\def\Width{\ensuremath{W}\xspace}
\def\Height{\ensuremath{H}\xspace}
\def\Channels{\ensuremath{C}\xspace}

\def\CodeRec{\ensuremath{\Code^{\text{rec}}}\xspace}

\def\WPCode{\ensuremath{\vb{w}}\xspace} %Maybe inconsistant use of bold vs caps?
\def\WPCodeK{\ensuremath{\WPCode_k}\xspace}
\def\WPCodeKAlign{\ensuremath{\WPCodeK^\text{align}}\xspace}
\def\WPCodeKRec{\ensuremath{\WPCodeK^\text{rec}}\xspace}

% Check usage
\def\WPCodeAlign{\ensuremath{\WPCode^\text{align}}\xspace}
\def\WPCodeRec{\ensuremath{\WPCode^\text{rec}}\xspace}

%Assert that NCourse + NDetails = 18
\def\NCoarse{\ensuremath{m}\xspace}
\def\NDetails{\ensuremath{(18-m)}\xspace}

\def\VGGActivations#1{\ensuremath{\widehat{\text{VGG}}^{{#1}}}}
\def\PerChannelWeightsOf#1{\ensuremath{w_{#1}^\text{pips}}}

\def\LossStructure{L_\StructureTensor}
\def\LossLPIPS{L_\text{PIPS}\xspace}
\def\LossMaskedLPIPS{L_\text{mask}\xspace}
\def\LossAlign{L_\text{align}}
\def\LossStyle{L_s}

\def\AppearanceRefImageK{\ensuremath{\RefImageK^{\text{app}}}\xspace}

\def\ActiveSegmentationRegionK{\rho_k}

	\begin{abstract}
		Seamlessly blending features from multiple images is extremely challenging
		because of complex relationships in lighting, geometry, and partial occlusion
		which cause coupling between different parts of the image. 
		Even though recent work on GANs enables synthesis of realistic hair or faces, it
		remains difficult to combine them into a single, coherent, and plausible image
		rather than a disjointed set of image patches.
		We present a novel solution to image blending, particularly for the problem of
		hairstyle transfer, based on GAN-inversion.
		We propose a novel latent space for image blending which is better at preserving
		detail and encoding spatial information, and
		propose a new GAN-embedding algorithm which is able to slightly modify images to
		conform to a common segmentation mask.
		Our novel representation enables the transfer of the visual properties from
		multiple reference images including specific details such as moles and wrinkles,
		and because we do image blending in a latent-space  we are able to synthesize
		images that are coherent. 
		Our approach avoids blending artifacts present in other approaches and finds a
		globally consistent image. 
		Our results demonstrate a significant improvement over the current state of the
		art in a user study, with users preferring our blending solution over 95 percent
		of the time. { \revision Source code for the new approach is available at
			\UrlOfRepo.
		}
	\end{abstract}

	%
	% The code below should be generated by the tool at
	% http://dl.acm.org/ccs.cfm
	% Please copy and paste the code instead of the example below.
	%
	
	% \ccsdesc{Computer systems organization~Robotics}
	% \ccsdesc[100]{Networks~Network reliability}
	
	%
	% End generated code
	%

	
	\begin{teaserfigure}
		\centering
		\includegraphics[width=\linewidth]
		{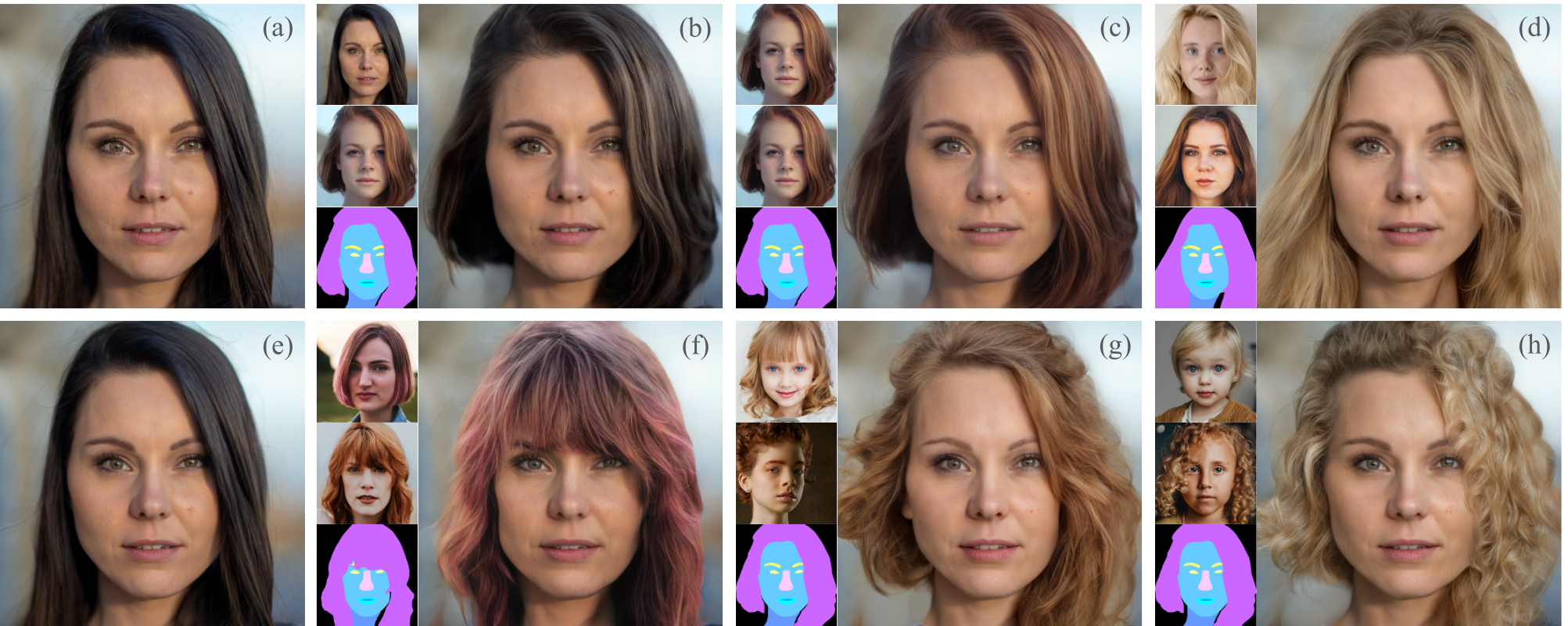}
		\caption{Hairstyle transfer is accomplished by transferring appearance (fine
			style attributes) and structure (coarse style attributes) from reference images
			into a composite image. In each inset, the appearance, structure, and target
			masks for a hairstyle image are shown on the left{\revision, with the hair shape
				in magenta}. Inset (a) is a reference image used for the face and background,
			and (e) is a reconstruction using our novel $FS$ latent space. In
			(b) a  reference image is used to transfer hair structure, but the hair's
			appearance is from the original face, and (c) transfers both appearance and
			structure from a hair reference, in
			(d) and (f) both structure and appearance attributes are transferred, (g) and
			(h) use a hair shape that is different from any of the reference images.}
		\label{fig:teaser}
	\end{teaserfigure}

	\maketitle
	
	\thispagestyle{empty}
    \pagestyle{plain} 
% 	\ifarXive
%   	   \thispagestyle{empty}
%   	   \pagestyle{plain} 
%   	\fi 
   	 
   	 \section{Introduction}
	
	Due to the rapid improvement of generative adversarial networks (GANs),
	GAN-based image editing has recently become a widely used tool in desktop
	applications for professional and social media photo editing tools for casual
	users. Of particular interest are tools to edit photographs of human faces. 
	In this paper, we propose new tools for image editing by mixing elements from
	multiple example images in order to make a composite image. Our focus is on the
	task of hair editing.

	% GANs are able to produce a wide variety of realistic synthetic images. 
	% According to different input conditions, there are two types of GANs, namely
	%conditional GANs and unconditional GANs.
	% Conditional GANs, e.g.~\cite{} trains the generative model for specific tasks,
	%and during inference, can directly edit the input image according to the new
	%conditions. The disadvantage of this method is that the generated image is not
	%as realistic as unconditional GANs, and unexpected artifacts may be produced.
	% On the other hand, unconditional GANs require GAN inversion (embedding) to
	%calculate the latent code for a given input image for further editing. Due to
	%the high visual quality of the generated images, most recent papers build on
	%\StyleGAN~\cite{} and \StyleGANTwo~\cite{}. By manipulating the latent code,
	%the generated image can be semantically edited.
	
	Despite the recent success of face editing based on latent space
	manipulation~\cite{abdal2019image2stylegan, abdal2020image2stylegan++,
		zhu2020improved}, most editing tasks operate on an image by changing global
	attributes such as \textit{pose}, \textit{expression}, \textit{gender}, or
	\textit{age}. 
	%achieving fine-grained manipulation of attributes while maintaining the realism
	%and fidelity
	%of unedited parts of the generated image is still an open challenge. 
	% In general, face editing by latent space manipulation has two main problems:
	%First, there may be obvious visual differences between the image after
	%embedding and the original image. Second, latent space manipulation usually
	%leads to changes in parts that do not need to be edited, such as the
	%background. 
	Another approach to image editing is to select features from reference images
	and mix them together to form a single, composite image. Examples of composite
	image editing that have seen recent progress are problems of hair-transfer and
	face-swapping.
	% Photo-bashing
	% Kit-bashing
	% Chimera 
	% Uses in-painting / image completion
	These tasks are extremely difficult for a variety of reasons. Chief among them
	is the fact that the visual properties of different parts of an image are not
	independent of each-other. The {\revii visual qualities} of hair, for example, are heavily
	influenced by ambient and reflected light as well as transmitted colors from the underlying face, clothing, and background. The pose of a head influences the {\reviii appearance of the }
nose, eyes and mouth, and the geometry of a person's head and
	shoulders influences shadows and the geometry of their hair. 
	Other challenges include disocclusion of the background {\revii (see the evaluation section Fig. 8 rows~1, 8 and Fig.~\ref{fig:qualitative-ablation-study})},  which happens when the 	hair region shrinks with respect to the background. Disocclusion of the face
	region can expose new parts of the face, such as ears, forehead, or the jawline {\revii (e.g. results are shown in Fig. 8 row 4 which exposes an ear)}.
	The shape of the hair is influenced by pose and also by the camera intrinsic
	parameters, and so the  pose might have to change to adapt to the hair.
	
	Failure to account for the global consistency of an image will lead to
	noticeable artifacts - the different regions of the image will appear
	disjointed, even if each part is synthesized with a high level of realism. In
	order for the composite image to seem plausible, our aim is to make a single
	coherent composite image that balances the fidelity of each region to the
	corresponding reference image while also synthesizing an overall convincing and
	highly realistic image.
	{\revii A key insight we present is that mixing images that are each of high quality, but also where each pixel has the same semantic meaning, produces new images with  fewer undesirable artifacts. One particular benefit of semantic alignment is that regions of the image which are disoccluded are filled with semantically correct image contents.  Therefore, we introduce a GAN-based semantic alignment step which generates high quality images similar to the input images, but which have a common semantic segmentation. When the semantic regions relevant to the task (e.g. hair) are aligned, artifacts caused by transparency, reflection, or interaction of the hair with the face, are less noticeable. This is illustrated in Fig.~\ref{fig:qualitative-ablation-study} which shows the artifacts that can occur when blending semantically dissimilar pixels. The value of alignment can also be seen in  Fig.~\ref{fig:method-overview}(d,g) in which the hair region of the identity image \ref{fig:method-overview}(b) is aligned to the mask, and blending between two semantically similar regions (e.g. the hair) is more likely to produce a plausible result than attempting to blend between semantically dissimilar regions. } 
	
	Previous methods of hair transfer based on GANs either use a complex pipeline of
	conditional GAN generators in which each condition module is specialized to
	represent, process, and convert reference inputs with different visual
	attributes~\cite{Tan_2020}, or make use of the latent space optimization with
	carefully designed loss and gradient orthogonalization to explicitly disentangle
	hair attributes~\cite{saha2021loho}.
	While both of these methods show very promising initial results, we found that
	they could be greatly improved. For example, both of them need pretrained
	inpainting networks to fill holes left over by misaligned hair masks, which may
	lead to blurry artifacts and unnatural boundaries. 
	We believe that better results can be achieved without an auxiliary inpainting
	network to fill the holes, as transitions between regions have higher quality if
	they are synthesized by a single GAN. 
	The previous methods do not make use of a semantic alignment step to
	merge semantic regions from different reference images in latent space, e.g. to
	align a hair region and a face region from different images.

	{\revii 
	The concepts of \textit{identity}, \textit{shape}, \textit{structure}, and \textit{appearance} were introduced in Michigan~\cite{Tan_2020} and then used by others including LOHO~\cite{saha2021loho} in order to describe different aspects of hair. The terms lack a precise definition, however \textit{appearance} broadly refers to the fine details (such as hair color) whereas \textit{structure} refers to coarser features (such as the form of locks of hair). The \textit{shape} of the hair is the binary segmentation region, and the \textit{identity} of a head-image encompasses all the features one would need to identify an individual. In this work, we propose Barbershop, a novel optimization method for
	photo-realistic hairstyle transfer, face swapping, and other composite image
	editing tasks applied to faces. Our approach, as illustrated in  Fig.~\ref{fig:teaser}, is capable of mixing together these four components in order to accomplish a variety of hair transfer tasks.}
	
\begin{figure}[thpb]
		\centering
		\includegraphics[width=\linewidth]{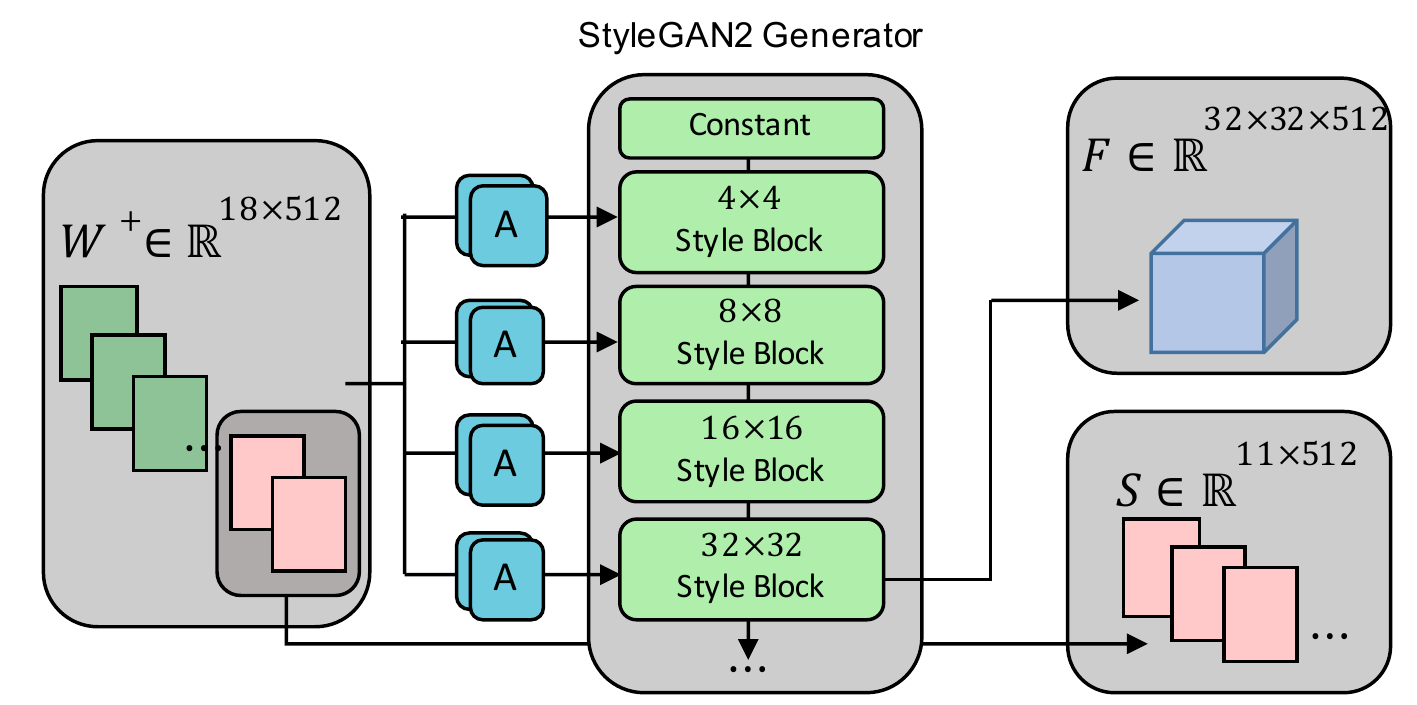}
		\caption{The {\revii relation between $FS$ and $W+$} latent space. The first \NCoarse (for \NCoarse=7) blocks of the $W+$ code are
			replaced by the output of style block \NCoarse to form a structure tensor
			$\StructureTensor$, and the remaining parts {\revii of $W+$} are used as an appearance code
			$\AppearanceCode$.}
		\label{fig:latent_space}
	\end{figure}

	Our approach uses GAN-inversion to generate 	high-fidelity reconstructions of reference images. We suggest a novel $FS$
	latent space which provides coarse control of the spatial locations of features
	via a \textit{structure tensor} $\StructureTensor$, as well as fine control of
	global style attributes via an \textit{appearance code} $S$. {\revision The
		elements of the novel space are illustrated in Fig.~\ref{fig:latent_space}}.
	This latent space allows a trade-off between a latent-code's capacity to
	maintain the spatial locations of features such as wrinkles and moles while also supporting latent code manipulation. We edit the codes to align reference images to  target feature locations. This {\revision semantic} alignment step is a
	key extension to existing GAN-embedding algorithms. It embeds images while at
	the same time slightly altering them to conform to a different segmentation
	mask. Then we find a blended latent code, by mixing reference images in a
	{\revision new spatially-aware} latent-space, rather than compositing images in
	the spatial domain. The result is a latent code of an image. By blending in the
	new 
	%spatially-aware 
	latent space, we avoid many of the artifacts of other image compositing
	approaches.

	Our proposed approach is demonstrated in Fig.~\ref{fig:teaser}. We are able to
	transfer only the \textit{shape} of {\revision a region corresponding to} the
	subject's hair (Fig.~\ref{fig:teaser}b).  We influence the shape by altering the hair region in a segmentation mask. We can also transfer the  structure
	(Fig.~\ref{fig:teaser}c) using the structure tensor, and the appearance (Fig.~\ref{fig:teaser}(d,~f)) by mixing appearance codes. Our approach also supports the use of different reference images
	to be used for structure vs the appearance code as shown in 
	Fig.~\ref{fig:teaser}(g,~h). 
	
	Our main contributions are:
	\begin{itemize}
		% Propigate to abstract and contributions
		% 1) New latent space. Better at preserving detail. Better to encode spatial
		%information.
		% 2) A new GAN-embedding algorithm for aligned embedding. Similar to previous
		%work, the algorithm can embed an image to be similar to an input image. In
		%addition, the image is slightly modified to conform to a new segmentation mask.
		% 3) A novel image compositing algorithm that can blend multiple images encoded
		%in our new latent space to yield a high quality result.
		
		\item A  novel latent space, called $FS$ space, for representing images. The
		new space is better at preserving details, and is more capable of encoding
		spatial information. 
		\item A new GAN-embedding algorithm for aligned embedding. Similar to previous
		work, the algorithm can embed an image to be similar to an input image. In
		addition, the image is slightly modified to conform to a new segmentation mask.
		\item  A novel image compositing algorithm that can blend multiple images
		encoded in our new latent space to yield a high quality results.
		\item We achieve a significant improvement in hair transfer, with our approach
		being preferred over existing state-of-the-art approaches by over 95\% of
		participants in a user study.
	\end{itemize}
	
	\section{Related Work}
	
	%\emph{Portrait Editing}
	\paragraph{GAN-based Image Generation.}
	Since their advent, GANs~\cite{goodfellow2014generative,
		radford2015unsupervised} have contributed to a surge in high quality image
	generation research. Several state-of-the-art GAN networks demonstrate
	significant improvements in the visual quality and diversity of the samples. 
	Some recent GANs such as ProGAN~\cite{karras2017progressive},
	\StyleGAN~\cite{STYLEGAN2018}, and \StyleGANTwo~\cite{Karras2019stylegan2} show
	the ability of GANs to produce very highly detailed and high fidelity images
	that are almost indistinguishable from real images. 
	Especially in the domain of human faces, these GAN architectures are able to
	produce unmatched quality and can then be applied to a downstream task such as
	image manipulation~\cite{shen2020interfacegan, abdal2019image2stylegan}. 
	StyleGAN-ada~\cite{Karras2020ada} showed that a GAN can be trained on limited
	data without compromising the generative ability of a GAN. 
	High quality image generation is also attributed to the availability of high
	quality datasets like FFHQ~\cite{STYLEGAN2018}, AFHQ~\cite{choi2020starganv2}
	and LSUN objects~\cite{yu15lsun}. 
	Such datasets provide both sufficient quality and diversity to train GANs and
	have further contributed to produce realistic applications. 
	On the other hand, BigGAN~\cite{brock2018large} can produce high quality samples
	using complex datasets like ImageNet~\cite{imagenet_cvpr09}. 
	Some other notable methods for generative modeling include Variational
	Autoencoders~(VAEs)~\cite{VAE2013}, PixelCNNs~\cite{Salimans2017PixeCNN},
	Normalizing Flows~\cite{chen2018neural}
	% How does this phrase ffit with the previous part of the sentence?
	and Transformer based VAEs~\cite{esser2020taming} also have some unique
	advantages.
	However, in this work, we focus on \StyleGANTwo trained on the FFHQ dataset
	because it is considered state of the art for face image generation. 
	
	\paragraph{Embedding Images into the GAN Latent Space}
	In order to edit real images, a given image needs to be projected into the GAN
	latent space. There are broadly two different ways to project/embed images into
	the latent space of a GAN. The first one is the optimization based approach.
	Particularly for \StyleGAN,  I2S~\cite{abdal2019image2stylegan} demonstrated
	high quality embeddings into the extended $W$ space, called $W+$ space, for real
	image editing. Several followup works~\cite{zhu2020domain,tewari2020pie} show
	that the embeddings can be improved by including new regularizers for the
	optimization. An Improved version of Image2StyleGAN
	(II2S)~\cite{zhu2020improved} demonstrated that regularization in $P-norm$ space
	can lead to better embeddings and editing quality. It is also noted that the
	research in these optimization based approaches with StyleGAN lead to commercial
	software such as Adobe Photoshop’s
	Neural Filters~\cite{Adobe}. The second approach in this domain is to use
	encoder based methods that train an encoder on the latent space. Some notable
	works~\cite{tov2021designing, richardson2020encoding} produce high quality image
	embeddings that can be manipulated. In this work, we propose several technical
	extensions to build on previous work in image embedding. 
	%adopt an improved optimization based embedding technique to achieve high
	%fidelity manipulations. % In particular, we make use of both, the activation
	%($F$ space see Section~\ref{sec:method}) as well as the W+ space of the
	%\StyleGANTwo for our embedding which leads to novel face swapping and hairstyle
	%manipulation applications.
	
	\paragraph{Latent Space Manipulation for Image Editing.} 
	GAN interpretability and GAN-based image manipulation  has been of recent
	interest to the GAN research community. There are broadly two spaces where
	semantic manipulation of images is possible: the latent and the activation
	space. Some notable works in the latent space manipulation domain try to
	understand the nature of the latent space of the GAN to extract meaningful
	directions for edits. For instance, GANspace~\cite{harkonen2020ganspace} is able
	to extract linear directions from the \StyleGAN latent space (W space) in an
	unsupervised fashion using Principal Component Analysis (PCA). Another notable
	work,  StyleRig~\cite{tewari2020stylerig} learns a mapping between a riggable
	face model and the \StyleGAN latent space. On the other hand, studying the
	non-linear nature of the \StyleGAN latent space, 
	StyleFlow~\cite{abdal2020styleflow} uses normalizing flows to model the latent
	space of \StyleGAN to produce various sequential edits. Another approach
	StyleCLIP~\cite{patashnik2021styleclip} uses text information to manipulate the
	latent space. The other set of papers focus on the layer
	activations~\cite{Bau:Ganpaint:2019, bau2020units} to produce fine-grained local
	edits to an image generated by \StyleGAN. Among them are
	TileGAN~\cite{fruhstuck2019tilegan},
	Image2StyleGAN++~\cite{abdal2020image2stylegan++},
	EditStyle~\cite{collins2020editing} which try to manipulate the activation maps
	directly to achieve a desired edit. Recently developed
	StyleSpace~\cite{wu2020stylespace} studies the style parameters of the channels
	to produce fine-grained edits. StylemapGAN~\cite{kim2021stylemapgan} on the
	other hand converts the latent codes into spatial maps that are interpretable
	and can be used for local editing of an image.
	
	\paragraph{Conditional GANs.}
	One of the main research areas enabling high quality image manipulation is the
	work on conditional GANs\linebreak (CGANs)~\cite{mirza2014conditional}. One way
	to incorporate a user's input for manipulation of images is to condition the
	generation on another image. Such networks can be trained in either
	paired~\cite{park2019SPADE,Zhu_2020} or unpaired
	fashion~\cite{Zhu_2017,zhu2017multimodal} using the cycle-consistency losses.  
	One important class of CGANs uses
	images as conditioning information. Methods such as pix2pix~\cite{pix2pix2017},
	BicycleGAN~\cite{zhu2017multimodal}, pix2pixHD~\cite{wang2018pix2pixHD},
	SPADE~\cite{park2019SPADE}, MaskGAN~\cite{fedus2018maskgan}, controllable person
	image synthesis~\cite{Men_2020},   SEAN~\cite{Zhu_2020} and
	SofGAN~\cite{chen2020free} are able to produce high quality images given the
	condition. 
	For instance, these networks can take a segmentation mask as an input and can
	generate the images consistent with manipulations done to the segmentation
	masks. 
	Particularly on faces, StarGANs1\&2~\cite{Choi_2018, choi2020starganv2} are able
	to modify multiple attributes. Other notable works,
	FaceShop~\cite{Portenier_2018}, Deep plastic surgery~\cite{Yang_2020},
	Interactive hair and beard synthesis~\cite{Olszewski_2020} and
	SC-FEGAN~\cite{Jo_2019} can modify the images using the strokes or scribbles on
	the semantic regions.
	For the hairstyle and appearance editing, we identified two notable relevant
	works. 
	MichiGAN~\cite{Tan_2020} demonstrated high quality hair editing using an
	inpainting network and mask-conditioned SPADE modules to draw new consistent
	hair. 
	LOHO~\cite{saha2021loho} decomposes the hair into perceptual structure,
	appearance, and style attributes and uses latent space optimization to infill
	missing hair structure details in latent space using the \StyleGANTwo generator.
	We compare with both these works quantitatively and qualitatively in
	Sec.~\ref{sec:competing}. 
	
	% % Table
	% \begin{table}%
	% \caption{Simulation Configuration}
	% \label{tab:one}
	% \begin{minipage}{\columnwidth}
	% \begin{center}
	% \begin{tabular}{ll}
	%   \toprule
	%   TERRAIN\footnote{This is a table footnote. This is a
	%     table footnote. This is a table footnote.}   & (200m$\times$200m) Square\\
	%\midrule
	%   Node Number     & 289\\
	%   Node Placement  & Uniform\\
	%   Application     & Many-to-Many/Gossip CBR Streams\\
	%   Payload Size    & 32 bytes\\
	%   Routing Layer   & GF\\
	%   MAC Layer       & CSMA/MMSN\\
	%   Radio Layer     & RADIO-ACCNOISE\\
	%   Radio Bandwidth & 250Kbps\\
	%   Radio Range     & 20m--45m\\
	%   \bottomrule
	% \end{tabular}
	% \end{center}
	% \bigskip\centering
	% \footnotesize\emph{Source:} This is a table
	%  sourcenote. This is a table sourcenote. This is a table
	%  sourcenote.
	
	%  \emph{Note:} This is a table footnote.
	% \end{minipage}
	% \end{table}%
	
	\section{Method}
	\label{sec:method}

	\subsection{Overview}
	
\begin{figure*}
		\centering
		
		\includegraphics[width=1.0\textwidth]{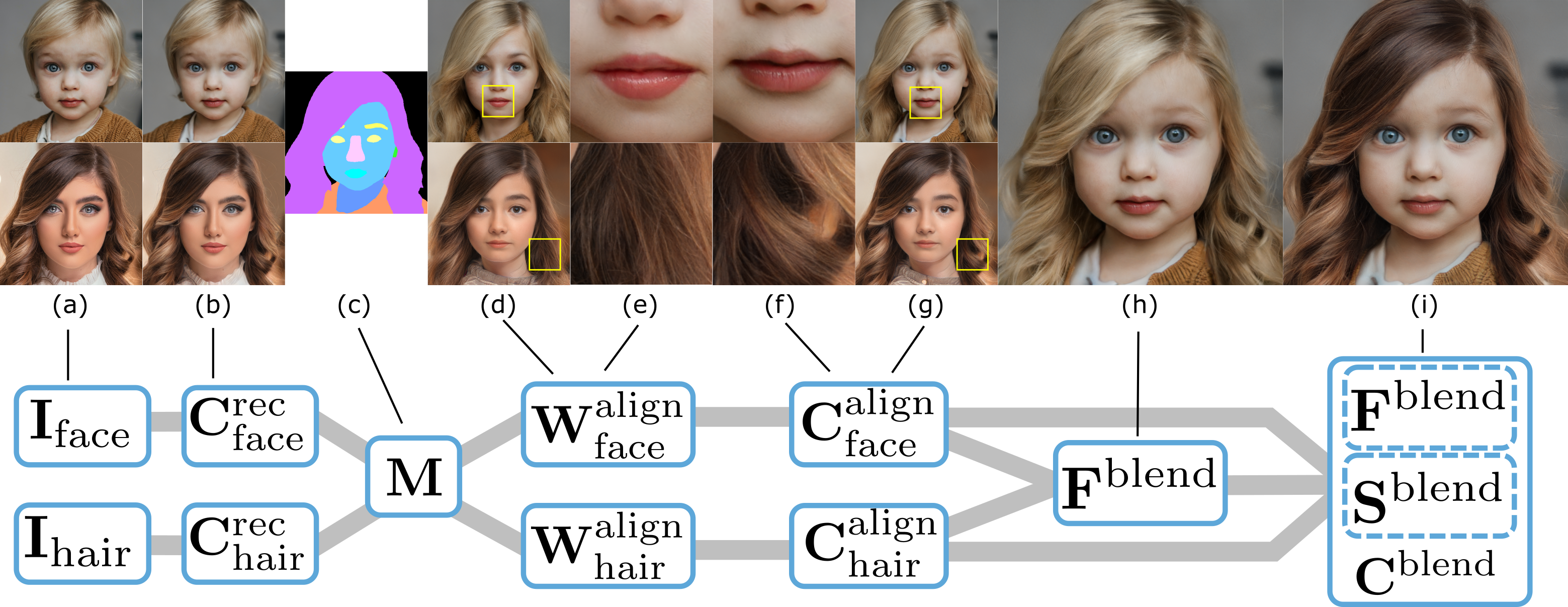}
		\caption{An overview of the method; (a) reference images for the face (top)
			and hair (bottom) features, (b) reconstructed images using the $FS$ latent
			space, (c) a target mask {\revii with hair region (magenta) from the hair image and all other regions from the face image}, (d) alignment in $W+$ space, (e) a close-up view of
			the face (top) and hair (bottom) in $W+$ space, (f) close-up views after details
			are transferred, (g) an entire image with details transferred, (h) the structure
			tensor is transferred into the blended image {\reviii but the appearance code is from $\AppearanceCode_\text{face}$}, and (i) the appearance code is
			optimized.  {\revision The data flow through our process is illustrated in the
				schematic at the bottom.} }
		\label{fig:method-overview}
	\end{figure*}

	We create composite images by selecting semantic regions 
	(such as hair, or facial features) from reference images and seamlessly blending
	them together.
	To this end, we employ automatic segmentation of reference images 
	and make use of a \textit{target} semantic segmentation mask image \Mask{}. 
	% Simply copying semantic regions from different reference images does not yield
	%a coherent new segmentation mask because some copied regions may intersect and
	%some regions in \Mask{} might not be covered. Therefore, the target mask
	%\Mask{} has to be adjusted automatically or edited manually to specify the
	%desired locations of each semantic region.
	% To train an automatic segmentation network and define semantic categories, we
	%use the CelebAMask-HQ~\cite{Lee2020} dataset. It defines 19 semantic categories
	%for face segmentation.
	To perform our most important example edit, hairstyle transfer, one can copy the
	hairstyle from one image, and use another image for all other semantic
	categories.
	More generally, a set of $K$ reference images, $\RefImageK$ for $k=1..K$, are
	each aligned to the target mask and then blended to form a novel image. 
	The output of our approach is a composite image, \BlendedImage, in which the region of
	semantic-category $k$ has the style of reference image $\RefImageK.$ See
	Fig.~\ref{fig:method-overview} for an overview.
	
	{\revii 
	Our approach to
	image blending finds a latent code for the blended image, which has the benefit
	of avoiding many of the traditional artifacts of image blending, particularly at
	the boundaries of the blended regions. 
	In particular, we build on the \StyleGANTwo architecture~\cite{Karras2019stylegan2}
	and extend the II2S~\cite{zhu2020improved} embedding algorithm.
% 	Our approach to image blending is based on \StyleGAN~\cite{Karras2019style,Karras2019stylegan2, Karras2020ada},  
% 	and \StyleGAN embedding algorithms to find	latent codes for given photographs, e.g.~\cite{abdal2019image2stylegan}.
	The II2S algorithm uses the inputs of the 18 affine style blocks of \StyleGANTwo
	as a single $W+$ latent code. The $W+$ latent code allows the input of each
	block to vary separately, but II2S is biased towards latent codes that have a
	higher probability according to the \StyleGANTwo training set. 
	There is a potential to suppress or reduce the prominence of less-common features in the training data. 
	}
	
	In order to increase the capacity of our embedding and capture image details, we
	embed images using a latent code 
	$\Code = (\StructureTensor, \AppearanceCode)$
	comprised of a \textit{structure tensor},
	$\StructureTensor\in\mathcal{R}^{32\times32\times512}$ 
	which replaces the output of the style block at layer $\NCoarse$ of the \StyleGANTwo image
	synthesis network, where $\NCoarse=7$ in our experiments, and an \textit{appearance code},
	$\AppearanceCode\in\mathcal{R}^{\NDetails \times512}$ that is used as input to the
	remaining style blocks. {\revision The relationship of our latent code to the
		StyleGAN2 architecture is illustrated in Fig.~\ref{fig:latent_space}.}  This
	proposed extension of {\reviii conventional} GAN embedding, which we call $FS$ space,
	provides more degrees of freedom to  capture individual facial details such as
	moles. However, it also requires a careful design of latent code manipulations,
	because it is easier to create artifacts.
	
	Our approach includes the following major steps:
	\begin{itemize}
		
		\item Reference images $\RefImageK$ are segmented and a \textit{target} segmentation is
		generated automatically, 
		or optionally the target segmentation is manually edited.  
		
		% R1:   Added this step as requested by Reviewer 1
		\item {\revision {\revii Embed input reference images $\RefImageK$ to find latent codes $\CodeKRec = (\StructureTensorKRec,\AppearanceCodeKRec)$,}} 

		\item {\revii Find latent codes
		$\CodeKAlign = (\StructureTensorKAlign, \AppearanceCodeKAlign)$ that are embeddings of images which match the target segmentation $\Mask$ while also being similar to the input images $\RefImageK$.} 
		
		\item A combined structure tensor $\StructureTensorBlend$ is formed by
		copying region $k$ of $\StructureTensorKAlign$ for each $k=1...K$. 
		
		\item Blending weights for the appearance codes
		$\AppearanceCodeKAlign$ are found so that the appearance code
		$\AppearanceCodeBlend$ is a mixture of the appearances of the aligned
		images.  The mixture weights are found using a novel masked-appearance loss
		function.
		
	\end{itemize}
	
	\subsection{Initial Segmentation} 
\begin{figure}
		\centering
		\includegraphics[width=0.8\columnwidth]{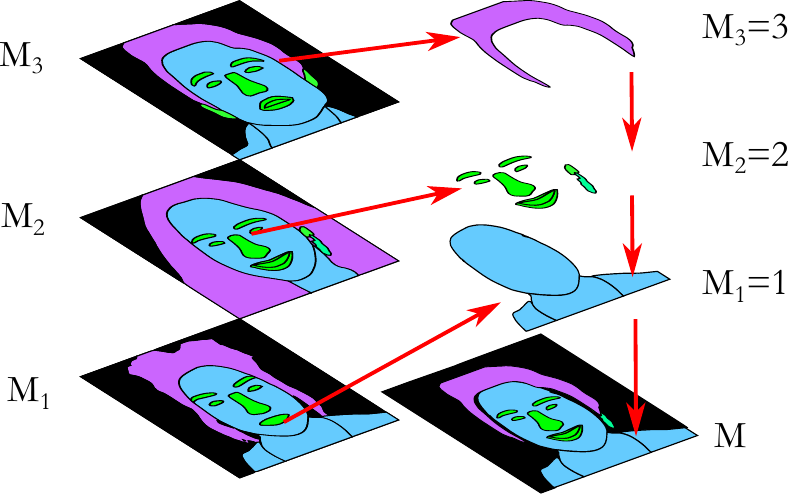}
		\caption{Generating the target mask. In this example, 19 semantic regions
			are relabeled to form four semantic categories including background. The label
			used in the target mask $\Mask$ is the largest index $k$ such that $\MaskK=k$.}
		\label{fig:make-mask}
	\end{figure}

	The first step is to select reference images, (automatically) segment them, and
	to select regions in the reference images that should be copied to the target
	image.
	Let  $\MaskK = \Segment(\RefImageK)$ indicate the segmentation of
	reference image $\RefImageK$,  where $\Segment$ is a segmentation network
	such as BiSeNET~\cite{Yu2018}.
	The aim is to form a composite image $\BlendedImage$ consistent with a \textit{target}
	segmentation mask $\Mask$ so that at locations $(x,y)$ in the image where $\Mask(x,y)=k$, the
	visual properties of $\BlendedImage$ will be transferred from reference images
	$\RefImageK$. The target mask $\Mask$ is created automatically, however one can
	also edit the segmentation mask manually to achieve more control over the shapes
	of each semantic region of the output.
	{\revision  Such editing may, for example, be useful to {\revii translate and scale} the regions
		in cases where the original images are not in the same pose. In the domain of
		cropped portrait images, images are coarsely aligned by placing the face region
		in the center of the image.} 
	In this exposition, we will focus our discussion on {\revision automatic
		processing (without editing)}. 
	To construct a target mask automatically, each pixel $\Mask(x,y)$ is set to a
	value $k$ that satisfies the condition that $\MaskK(x,y) = k$. {\revii To resolve conflicts between segmentation masks (the condition $\MaskK(x,y) = k$ is satisfied for two or more $k$), we assume that the values $k$ are sorted according to priority, so that higher values of $k$ are composited over lower values of $k$.
	%In this context, it is also useful to relabel segmentation images so that regions that are not important to the compositing task (such as left vs right eyes) are given the same value.
	
	A conflict would happen, for example, if a pixel is covered by \textit{skin} (label
	1) in a reference image corresponding to the label \textit{skin}, but also
	covered by \textit{hair} (label 13) in a reference image corresponding to
	\textit{hair}, and so the label for \textit{hair} would be chosen.  
	Some pixels may not be covered by any of the segmentation masks (the condition $\MaskK(x,y) = k$ is not satisfied for any $k$).
	}
	In this case, a portion of the target mask
	will be in-painted using a heuristic method {\revii explained in the supplementary materials}. 
	The process of automatically creating a mask is illustrated in
	Fig.~\ref{fig:make-mask}.
	%Regions of $\Mask$ that remain uncovered are marked with a special
	%\textit{don't care} value.
	
	% Use heuristics to fill in the missing regions, discuss them in limitation?

	\subsection{Embedding:}
	Before blending images, we first align each image to the target mask $\Mask$. 
	This is important because the appearance of many features such as hair, nose,
	eyes, and ears depend on the pose of the head as a whole, which introduces a
	dependency between them.   
	Our approach to aligning the reference images has two parts:
	\begin{enumerate}
		\item \textbf{Reconstruction:} A latent code $\CodeRec_k$ is found to
		reconstruct the input image $\RefImageK$. 
		\item \textbf{Alignment:} A nearby latent code $\Code_k^\text{align}$ is found
		that minimizes the cross-entropy between the generated image and the target mask
		$\Mask$. 
	\end{enumerate}
	
	\subsubsection{Reconstruction} \label{sec:reconstruction}
	Given an image $\RefImageK$ we aim to find a code $\CodeKRec$ so that
	$\Generator(\CodeKRec)$ reconstructs the image $\RefImageK$, where $\Generator$ is the \StyleGANTwo
	image synthesis network. 
	Our approach to finding a reconstruction code $\CodeKRec$ is to initialize it
	using II2S~\cite{zhu2020improved}, 
	which finds a latent code $\WPCode_k^\text{rec}$ in the $W+$ latent-space of
	\StyleGANTwo.
	The challenge of any reconstruction algorithm is to find a meaningful trade-off
	between reconstruction quality and suitability for editing or image compositing.
	{\revii 
	The $W$ latent space of \StyleGANTwo{} has only 512 components. 
	While it is expressive enough to capture generic details, such as wrinkles, it is not possible to encode specific details of a particular face (such as the precise locations of moles, wrinkles, or
	eyelashes). 
	The use of $W+$ space and II2S improves the expressiveness of the
	latent space, but it is still not sufficient for a faithful reconstruction.
	}
	
\begin{figure}
		\centering
		\includegraphics[width=0.99\linewidth]{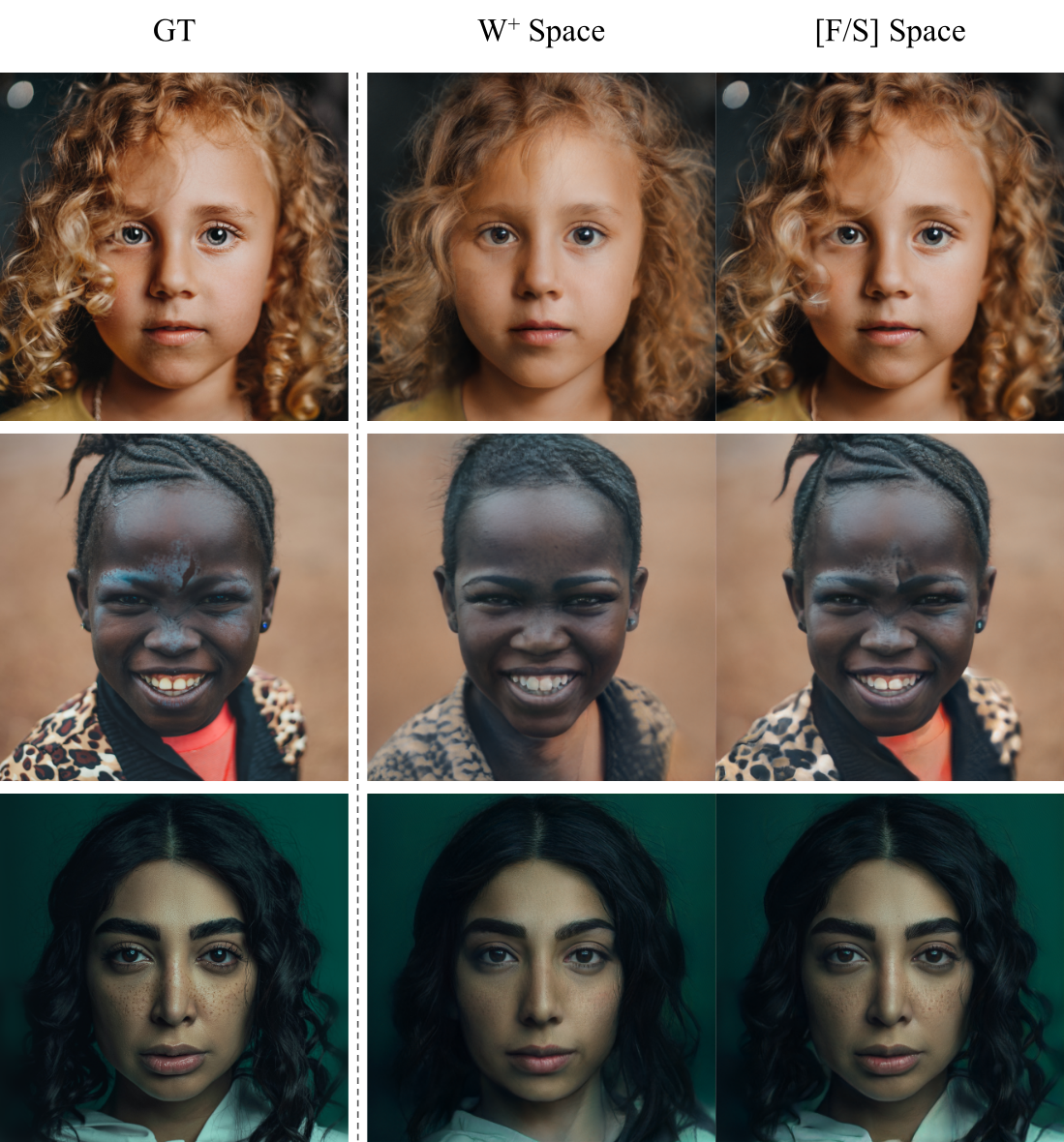}
		\caption{Reconstruction results on different spaces; (top row) in $W+$
			space, structure of the subject's curly hair on the left of the image is lost,
			and a wisp of hair on her forehead as well as her necklace is removed, but they
			are preserved in $FS$ space;  (middle row) the hair and brow furrows details are
			important to the expression of the subject, they are not preserved in $W+$ space
			but they are in $FS$ space; (bottom row) the ground-truth image has freckles,
			without noise optimization this is not captured in $W+$ space but it is
			preserved in $FS$ space. }
		\label{fig:reconstruction}
	\end{figure}

	One possible approach is noise embedding that leads to embedded images
	with almost perfect reconstruction, but leads to strong overfitting which
	manifests itself in image artifacts in downstream editing and compositing tasks.
	Our idea is to embed into a new latent space, called $FS$ space, that provides
	better control than $W+$ space without the problems of noise embedding.
	Similarly to $W+$ embedding, we need to carefully design our compositing
	operation, so that image artifacts do not manifest themselves.
	The difference between reconstruction in $W+$ vs $FS$ space is shown in
	Fig.~\ref{fig:reconstruction}, illustrating that key identifying features of a
	person (such as a facial mole)  or important characteristics of a subject's
	expression (hairstyle, furrows in the brow) are captured in the new latent
	space. 
	
	%{\color{red} NEEDS A FIGURE TO SHOW THE DIFFERENCE BETWEEN SPECIFIC AND GENERIC
	%EMBEDDING}
	
	% In \StyleGANTwo, details are added to an image using random "noise" vectors
	%which are added after each style block, however treating these details as
	%random noise can change the identifying marks on a character.
	% Treating noise as part of a latent code is also problematic because, unlike
	%style codes, the noise signal is spatially correlated.
	We capture specific facial details by using a spatially correlated signal as
	part of our latent code.
	We use the output of one of the style-blocks of the generator as a
	spatially-correlated \textit{structure-tensor} $\StructureTensor$, which
	replaces the corresponding blocks of the $W+$ latent. The choice of a particular
	style block is a design decision, however each choice results in a
	different-sized latent code and in order to keep the exposition concise our
	discussion will use style-block eight. 
	
	The resulting latent code has more capacity than the $W+$ latent-codes, and we
	use gradient descent initialized by a $W+$ latent-code in order to reconstruct each
	reference image. We form an initial structure tensor
	$\StructureTensorKInit=\Generator_\NCoarse(\WPCodeKRec)$, 
	and the remaining \NDetails blocks of $\WPCode_k^\text{rec}$ are used to initialize the
	appearance code $\AppearanceCode^\text{init}_k$. 
	Then we set $\CodeKRec$ to the nearest local minimum of 
	\begin{align} \CodeKRec &=\arg\min_\Code  \LossLPIPS(\Code)  +
		\LossStructure .\label{eq:loss-rec}\\
		\intertext{where}
		\LossStructure &= \|\StructureTensor-\StructureTensor^{\text{init}}_k\|^2 \label{eq:l-f}
	\end{align}
	The term $\LossStructure$ in the loss function~\eqref{eq:l-f} encourages
	solutions in which $\StructureTensor$ remains similar to the activations of a
	$W+$ code so that the result remains close to the valid region of the
	\StyleGANTwo latent space.
	%This regularization term is consistent with the prior that noise in the
	%generator is zero-mean with small variance.  

	\subsubsection{Alignment}

	We now have each reference image $\RefImageK$ encoded as a latent code
	$\CodeKRec$ consisting of a tensor $\StructureTensorKRec$ and
	appearance code $\AppearanceCodeKRec$.
	While $\CodeKRec$ captures the appearance of the reference image
	$\RefImageK$, the details will not be aligned to the target segmentation. 
	Therefore, we find latent codes $\CodeKAlign$ that match the target
	segmentation, and which are nearby  $\CodeKRec$ {\revii in latent space}.
	We find that directly optimizing $\CodeKAlign$ is challenging because the
	details of $\StructureTensorKRec$ are spatially correlated. Instead we
	first search for a $W+$ latent code, $\WPCodeAlign$ for the aligned
	image and then we transfer details from $\StructureTensorKRec$ into
	$\StructureTensorKAlign$ where it is safe to do so.  {\revision Results
		of aligning images to the target mask are shown in
		Fig.~\ref{fig:method-overview} and Fig.~\ref{fig:qualitative-ablation-study}.}
%
% 	We build on the idea that we can retrieve an image from GAN latent space that
% 	conforms to a segmentation mask $\Mask$ using the cross-entropy loss of the
% 	given segmentation mask and a segmentation mask derived from the GAN output
% 	using a pre-trained segmentation network. However, here we deal with a
% 	specialized version of this problem. We would like to retrieve a latent
% 	representation that conforms to a segmentation mask and that is similar to a
% 	given reference image in a given region. Simply initializing with the input
% 	representation and then optimizing for a segmentation loss does not work. The
% 	image would not be similar enough to the input image.

	{\revii Our approach for finding a latent code is to compose the generator $\Generator$ with a semantic segmentation network $\Segment$ to construct the differentiable function $\Segment\circ\Generator$, which is a differentiable generator of semantic segmentations. Using GAN inversion (e.g. II2S) on this new generator to minimize an appropriate loss such as cross-entropy, it is possible to find a latent code $\WPCodeAlign$, so that $\Generator(\WPCodeAlign)$ is an image whose segmentation matches the target segmentation. However, GAN inversion is  ill-posed for segmentation masks, as many images could produce the same semantic segmentation. We also aim to find an image that is also as similar as possible to the original latent code of a reference image. We therefore experimented 	with a combination of $L_2$, $L_1$, and style losses to preserve the content of the reference images and found that only using the style loss, produces the best
	results.}

	In order to preserve the style between an aligned image
	$G(\WPCodeAlign)$ and the original image $\RefImageK$, we use a masked
	style-loss. 
	The masked loss described in LOHO~\cite{saha2021loho} uses a static mask in
	order to compute the gram matrix of feature activations only within a specific
	region, whereas each step of gradient descent in our method produces a new
	latent code, and leads to a new generated image and segmentation. Therefore the
	mask used at each step is dynamic. 
	Following~\cite{saha2021loho}, we base the loss on the gram matrix 
	\begin{align} 
		\mathbf{K}_{\ell}(\Image) = \gamma_{\ell}^T \gamma_{\ell}^{},
		\label{eq:gramm-of-layer}
	\end{align}
	where $\gamma_\ell^{}\in\mathcal{R}^{H_\ell \Width_\ell\times C_\ell}$ is a
	matrix formed by the activations of layer $\ell$ of the VGG network. In
	addition, we define a region mask for region $k$ of an image $\Image$ as  
	\begin{align}
		\ActiveSegmentationRegionK(\Image)=1\{\textsc{Segment}(\Image)=k\}, 
	\end{align}
	where $1\{\cdot\}$ is the indicator function, so $\ActiveSegmentationRegionK$ is an indicator for the
	region of an image that is of semantic category $k$. Then the style loss is the
	magnitude of the difference between the gram matrices of the images generated by
	a latent code $\WPCode$ and the target image $\RefImageK$, and it is evaluated
	only within semantic region $k$ of each image
	\begin{align}
		\LossStyle &= \sum_{\ell} \|\mathbf{K}_{\ell}(\ActiveSegmentationRegionK(\Generator(\WPCode))\odot
		\Generator(\WPCode))-\mathbf{K}_{\ell}(\ActiveSegmentationRegionK(\RefImageK)\odot \RefImageK))\|^2,
	\end{align}
	where the summation is over layers  $relu1\_2$, $relu2\_2$, $relu3\_3$, and
	$relu4\_3$ of VGG-16, as was done in LOHO~\cite{saha2021loho}. The formulation 
	$\ActiveSegmentationRegionK(\RefImageK)\odot \RefImageK$ describes the masking of an image by setting all
	pixels outside the semantic region $k$ to 0. 
	
	In order to find an aligned latent code, we use gradient descent to minimize a
	loss function which combines the cross-entropy of the segmented image, and the
	style loss
	\begin{align}
		\LossAlign(W) = \textsc{XEnt}(\Mask, \textsc{Segment}(G(W))) + \lambda_s
		\LossStyle, \label{eq:l-align} 
	\end{align}
	where $\textsc{XEnt}$ is the multi-class cross-entropy function. 
	
	{\revii We optimize only the portion of  $W+$ space that corresponds to $\StructureTensor$ during alignment because the goal is to construct $\StructureTensor^\text{align}$.}
	We rely on early-stopping to keep the $\WPCodeAlign$ latent code nearby
	the initial reconstruction code  $\WPCodeRec,$  and $\lambda_s$ is set to the value
	recommended by~\cite{saha2021loho}. {\revision We stop at 100 iterations, but
		find that between 50 and 200 iterations produce qualitatively similar results.}
	% R5 -- What number of iterations for ealrly stopping???
	
	\subsubsection{{\revision Structure Transfer}} 
	{\revision Alignment using  $\WPCodeAlign$ produces plausible images but
		some details are changed as shown in Fig. \ref{fig:method-overview}(e) vs Fig.
		\ref{fig:method-overview}(f). In cropped portrait images, regions which overlap
		are often spatially aligned, and so we transfer the structure of the
		reconstructed images within those regions.} {\revii Note that the target mask is not always perfectly aligned to regions in the reference images, due to inpainting or, as shown in Fig.\ref{fig:hair-gallery}, if the hair shape comes from a third reference image. }
	In order to transfer the structure and appearance from image $\RefImageK$ into
	$\StructureTensorK$, we use binary masks to define safe regions to copy
	details,  
	\begin{align}
		\alpha_k(x,y) & =1\{\Mask(x,y)=k\}, \label{eq:alpha_k} \\
		\beta_k(x,y) & =1\{\MaskK(x,y)=k\}, \label{eq:beta_k} 
	\end{align}
	where $1\{\cdot\}$ is the indicator function. 
	{\reviii Let $\alpha_{k,\ell}$ denote $\alpha_k$  downsampled using bicubic-resampling to match
	the dimensions of the activations in layer $\ell$,  noting that the resampled mask is no longer binary. 
	The mask $\beta_{k, \ell}$ is similarly defined. At layer $m$,}  
	the mask $\alpha_{k,\NCoarse}\cdot\beta_{k,\NCoarse}$ is a {\revision soft} region {\revision
		with membership values in the range $[0,1]$ }where it is safe to copy structure
	from the code 
	$\StructureTensorKRec$ because the semantic classes of the target and
	reference image are the same.   
	The mask $(1-\alpha_{k,\NCoarse}\cdot\beta_{k,\NCoarse})$ is a {\revision soft} region where
	we must fall-back to $\WPCodeKAlign$, which has less capacity to
	reconstruct detailed features.  {\revision These values are not restricted to
		binary values because they are the result of bicubic downsampling.}
	%We omit the target size, as it can be determined from context.   
	We use the structure-tensor 
	\begin{align}
		\StructureTensorKAlign &= \alpha_{k,\NCoarse}\cdot\beta_{k,\NCoarse}\cdot
		\StructureTensorKRec  + (1-\alpha_{k,\NCoarse}\cdot\beta_{k,\NCoarse})\cdot 
		\Generator_\NCoarse(\WPCodeKAlign), 	\label{eq:f-align}
	\end{align}
	where $\Generator_\NCoarse(\WPCodeKAlign)$ is the output of style-block \NCoarse of the
	generator applied to input $\WPCodeKAlign$. 
	We now have an aligned latent representation $\CodeKAlign$ for each
	reference image $k$. Next we can composite the final image by blending the
	structure tensors 
	$\StructureTensorKAlign$ and appearance codes
	$\AppearanceCodeKAlign$ as described in the next two subsections.
	
	\subsection{Structure Blending:}
	In order to create a blended image, we combine the structure tensor elements of
	$\CodeKAlign$ using weights $\alpha_{k,\NCoarse}$ to mix the structure tensors, so
	\begin{align}
		\StructureTensorBlend &=\sum_{k=1}^K \alpha_{k,\NCoarse} \odot \StructureTensorKAlign.
	\end{align}
	The coarse structure of each reference image can be composited simply by
	combining the regions of each structure tensor, however mixing the appearance
	codes requires more care.

	\subsection{Appearance Blending}
	
	Our approach to image blending is to find a single style code
	$\AppearanceCodeBlend$, which is  a mixture of the $K$ different
	reference codes $\AppearanceCodeK, k=1..K$.
	To find $\AppearanceCodeBlend$ we optimize a \textit{masked} version of
	the LPIPS distance function as a loss. 
	Following {\revision the notation of}~\cite{Zhang2018}, we will describe a \textit{masked} version of LPIPS that will be used to solve for $\AppearanceCodeBlend$.  
	
	{\revii 
	A complete motivation of LPIPS is beyond the scope of this work, we refer the reader to~\cite{Zhang2018} and the \textit{lin} version of $\LossLPIPS$ for details. The focus of this section is to explain our modification, which extends it to be used to compare $K$ different masked-images. 
	First, we include the formula for the original $\LossLPIPS$ function for comparison. Let $\VGGActivations{\ell}$ indicate the activations of layer $\ell$ of convnet (VGG) normalized across the channel-dimension as described in~\cite{Zhang2018}.
	The shape of that tensor is $\Width_\ell, \Height_\ell,$ and it has $\Channels_\ell$ channels. 
	To compare an image $\Image$ to another image $\Image_o$
	\begin{align}
		\LossLPIPS &= \sum_{\ell} \frac{1}{\Height_\ell
			\Width_\ell}
			\sum_{i,j}\!\|
			\PerChannelWeightsOf{\ell} \odot 
    			\Delta^{\ell}_{i,j}(\Image, \Image_o)\|^2
	\end{align}
	{where}
	\begin{align}
		\Delta^{\ell}_{i,j}(\Image, \Image_o)&=\VGGActivations{\ell}_{i,j}(\Image)
    			-\VGGActivations{\ell}_{i,j}(\Image_o))
	\end{align}
    and the vector $\PerChannelWeightsOf{\ell}$  is a learned vector of per-channel weights associated with layer $\ell$.   Similar to~\cite{Zhang2018}, we do not use every layer of VGG and instead set $\ell$ to only the three layers (\texttt{conv1}-\texttt{conv3}) of VGG. 
   
	A \textit{masked} version of the loss uses the masks $\alpha_{k,\ell}$ to compare a region $k$ of an arbitrary image $\Image$ with each of the $k$ corresponding aligned images $\AlignedRefImageK=\Generator(\CodeKAlign)$, so 
	
	\begin{align}
		\LossMaskedLPIPS\!&=\!\sum_{\ell} 
		\frac{1}{\Height_\ell\Width_\ell}\!
		\sum_{kij}
			(\alpha_{k\ell})_{ij} \|\PerChannelWeightsOf{\ell} \odot 
			\Delta^{\ell}_{i,j}(\Image, \AlignedRefImageK)
		\|^2, \label{eq:l-masked}
	\end{align}
	 where  $\Delta$ is defined as above and $\alpha_{k, \ell}$ is a mask which has been resampled {\revision(bicubic)} to match the dimensions  of each layer.  
	
	When solving for $\AppearanceCodeBlend$, we want to constrain the latent-code so that it stays within a region of latent space that contains the aligned reference codes ($\AppearanceCodeKAlign$). Due to the ill-conditioned nature of GAN inversion, an unconstrained solution will tend to overfit the loss and find an appearance code that is arbitrarily far from any of the inputs. A constrained solution restricts the set of possible latent-codes to a small portion of the embedding space.  Our approach is to find a set of $k$
	different blending weights $U = \{\vb{u}_k\}$ so that each $\vb{u}_k$ is a vector in  $\mathcal{R}^{\NDetails\times512}$.
	The blended code $\AppearanceCodeBlend$ satisfies
	\begin{align}
		\AppearanceCodeBlend &= \sum_k \vb{u}_k \odot \AppearanceCodeKAlign \label{eq:s-blend} 
	\end{align}
	and the weights satisfy the constraints 
	\begin{align}
	    \sum_k \vb{u}_k &= \vb{1}, & \vb{u}_k &\geq 0 \label{eq:convexity}
	\end{align}
    so that each element of $\AppearanceCodeBlend$ is a convex combination
	of the corresponding elements in reference codes $\AppearanceCodeKAlign$. 
	We find $\CodeBlend$ using projected gradient
	descent~\cite{landweber1951iteration} to minimize $\LossMaskedLPIPS$ with $I=G(\CodeBlend)$, and \CodeBlend itself is a function {\reviii of $U$} in equation \eqref{eq:s-blend}.  We initialize $U$  so that the blended image would be a copy of one of the 	reference images  and solve for the values that minimize $\LossMaskedLPIPS$ subject to the constraints \eqref{eq:convexity}.
	}
	\subsection{Mixing Shape, Structure, And
		Appearance}\label{sec:mixing-shape-structure-and-apperance}
	
	We have presented an approach to create composite  images using a set of
	reference images $\RefImageK$ in which we transfer the shape of a region, the
	structure tensor information $\StructureTensorK$, and also the appearance
	information $\AppearanceCodeK$. The LOHO~\cite{saha2021loho} approach
	demonstrated that different reference images can be used for each attribute
	(shape, structure, and appearance) and our approach is capable of doing the
	same. We simply use an additional set of images $\AppearanceRefImageK$ for
	the appearance information, and we set $\AppearanceCodeK$ using the last \NDetails
	blocks of the $W+$ code that reconstructs  $\AppearanceRefImageK$ instead
	of using the latent code that reconstructs $\RefImageK$. The additional images 
	$\AppearanceRefImageK$ do not need to be aligned to the target mask. We	show example of mixing shape, structure, and appearance in Fig.~\ref{fig:teaser}(g,h). The larger structures of the hair (locks of hair,
	curls)  are transferred from the structure reference, and the hair color and
	micro textures are transferred from  the appearance image.

	\begin{figure*}
		\centering
		\includegraphics[width=0.99\linewidth]{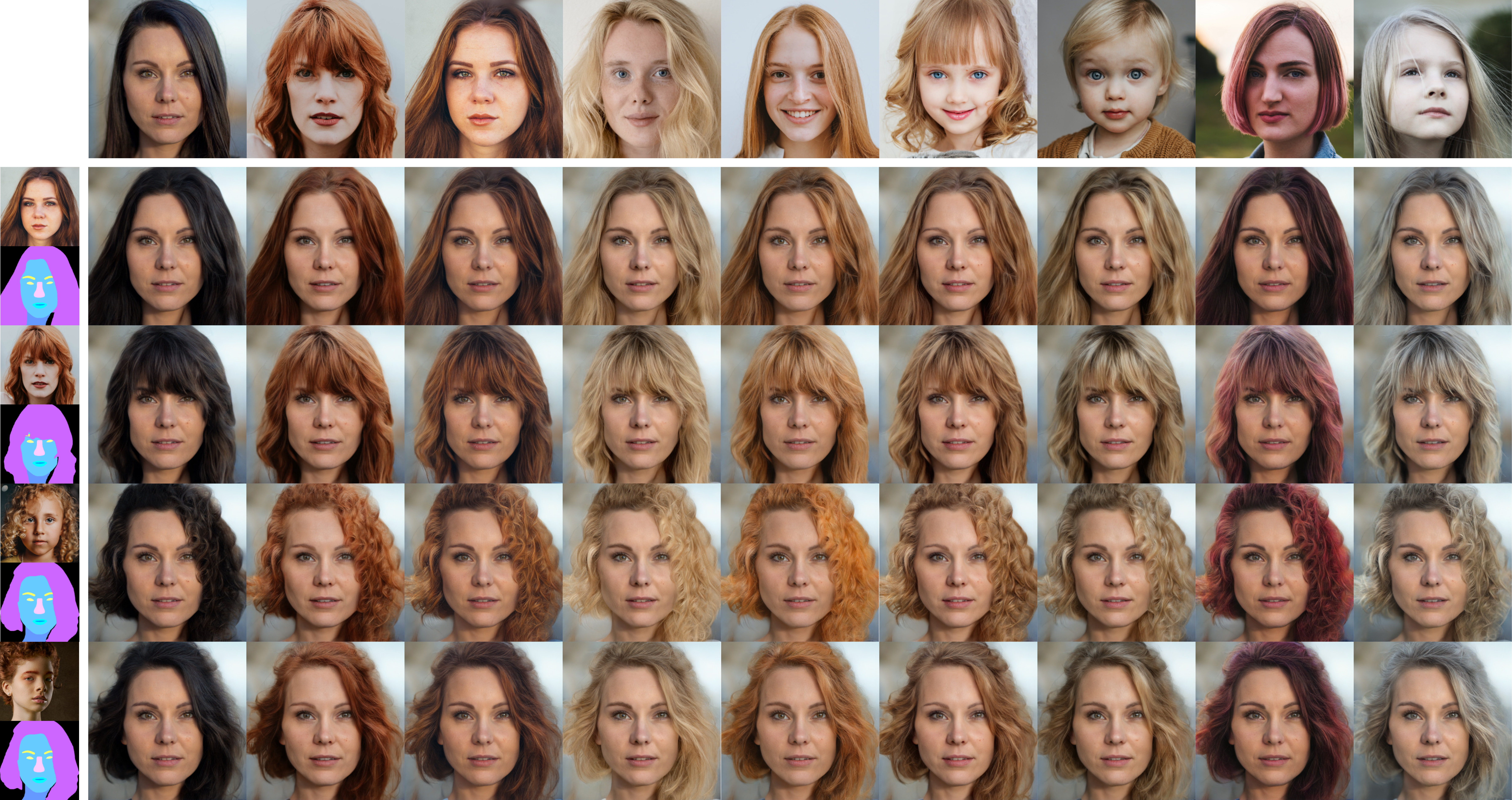}
		\caption{Hair style gallery showing different hairstyles applied to a person
			by varying the hair structure and appearance. Reference images for the hair
			appearance are shown at the top of each column, Reference images for the hair
			structure and the target segmentation masks are shown to the left of each row.
			Also note that in the last two rows, the hair shape is different from the hair
			shape of the structure reference images.}
		\label{fig:hair-gallery}
	\end{figure*}

	\begin{figure*}
		\centering
		\includegraphics[width=0.99\linewidth]{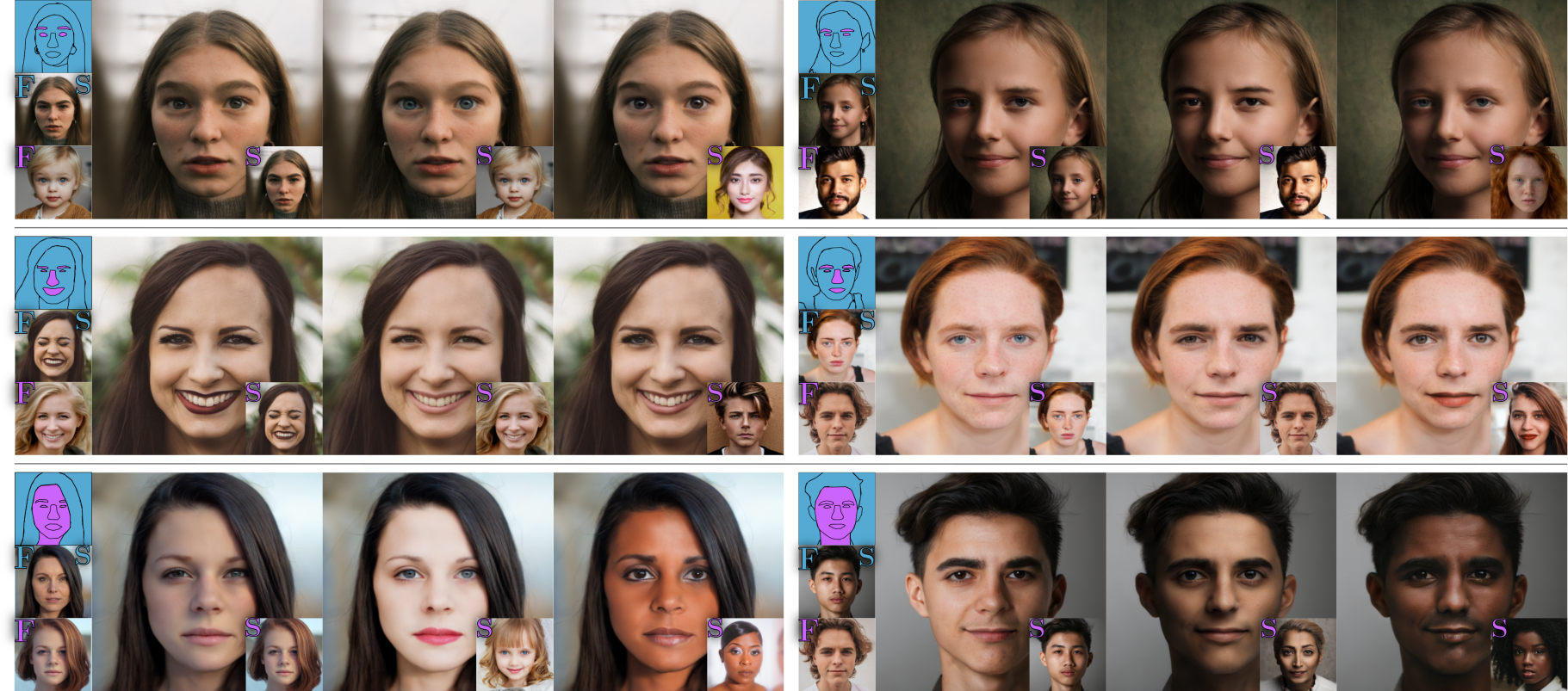}
		\caption{Face swapping results achieved by our method. Each example shows
			smaller insets: a target segmentation mask (top left) with the source regions indicated using cyan and magenta, an 'identity' image (center left), corresponding to the cyan region which includes all regions except the ones being transferred,  a structure reference image (bottom left) and an appearance image (bottom right) each used to transfer the structure or appearance of the magenta regions. 
			The first row shows examples of eye and eyebrow transfer by varying the appearance
			reference images; the second row shows examples of eye, eyebrows, nose, and mouth
			transfer; the third row shows examples transferring the entire facial region including skin.}
		\label{fig:face-swapping}
	\end{figure*}

	\section{Results}\label{sec:results}
	
	In this section, we will show a quantitative and qualitative evaluation of our
	method. 
	We implemented our algorithm using PyTorch and a single NVIDIA TITAN Xp graphics
	card.
	The process of finding an II2S embedding takes 2 minutes per image on average,
	the optimization in \eqref{eq:loss-rec} takes 1 minute per image. The resulting
	codes are saved and reused when creating composite images. For each composite
	image, we solve equation \eqref{eq:l-align} and then \eqref{eq:l-masked} to
	generate a composite image in an average time of two minutes. 
	
	\subsection{Dataset}
	% For the future -- I think it would have been smarter to try to use an existing
	%dataset
	% -- FFHQ has facial landmarks (for aligning faces).....
	We use a set of 120 high resolution ($1024\times 1024$) images
	from~\cite{zhu2020improved}.
	From these images, 198 pairs of images were selected for the hairstyle transfer
	experiments based on the variety of appearances and hair shape. 
	Images are segmented and the \textit{target} segmentation masks are generated
	automatically.

	\subsection{Competing methods}\label{sec:competing}
	We evaluate our method by comparing the following three algorithms:
	MichiGAN~\cite{Tan_2020}, LOHO~\cite{saha2021loho}, and our proposed method. 
	
	The authors of LOHO and MichiGAN provide public implementations, which we used
	in our comparison. 
	However, MichiGAN uses a proprietary inpainting module that the authors could
	not share. The authors supported our comparison by providing some inpainting
	results for selected images on request.
	LOHO also uses a pretrained inpainting network. Based on our analysis, both
	methods can be improved by using different inpainting networks as proposed in
	the initial papers. We therefore replaced both inpainting networks by the
	current state of the art CoModGAN~\cite{zhao2021comodgan} trained on the same
	dataset as LOHO.
	All hyperparameters and configuration  options were kept at their default
	values. 
	
	Our approach was used to reconstruct images using a fixed number of gradient
	descent iterations for each step. To solve for $\CodeKRec$ in equation
	\eqref{eq:loss-rec} we used 400 iterations, to solve for $\Code_k^\text{align}$
	using  \eqref{eq:l-align} we stopped after 100 iterations, and to solve for the
	blending weights $u$ using  \eqref{eq:l-masked} we stopped after 600 iterations.
	
	\textbf{\revision Source code for our method will be made public after an
		eventual publication of the paper at \UrlOfRepo.}

	\subsection{Comparison}
	
	\subsubsection{User Study}\label{sec:user-study}
\begin{figure*}[htbp]
		\centering
		\includegraphics[width=0.95\linewidth]{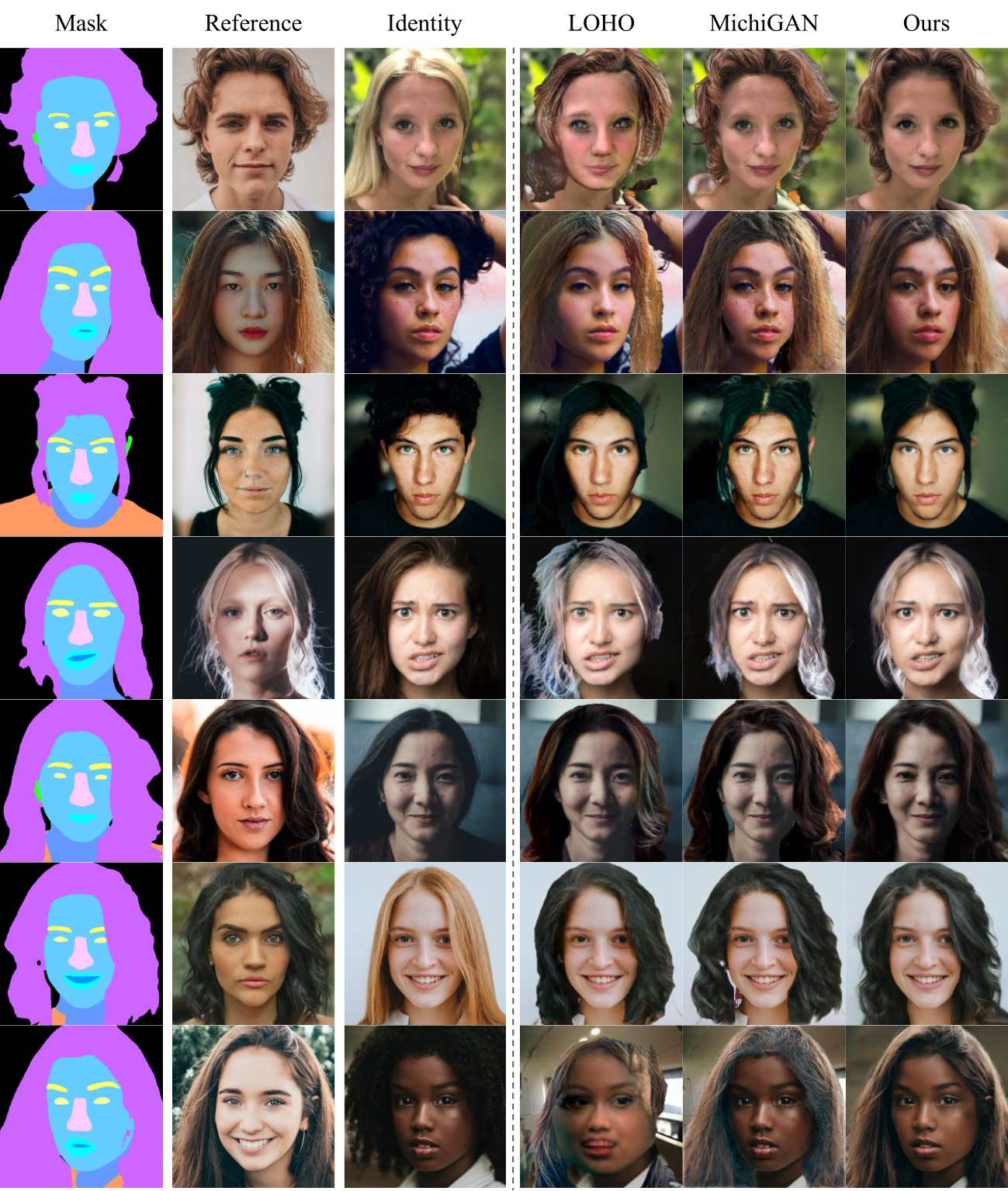}
		\caption{Comparison of our framework with two state of the art methods: LOHO
			and MichiGAN. Our results show improved  transitions between hair  and other
			regions, fewer disocclusion artifacts, and a better consistent handling of
			global aspects such as lighting.
		}
		\label{fig:comparison}
	\end{figure*}

	We conducted a user study using Amazon's Mechanical Turk to evaluate the
	hairstyle transfer task. For this task we use the 19-category segmentation from
	CelebAMask-HQ. A \textit{hairstyle} image was used as the reference for the the
	corresponding category in CelebAMask-HQ, and an \textit{Identity} image was used
	for all other semantic categories. We generated composite images using our
	complete approach and compared the results to LOHO~\cite{saha2021loho} and to
	MichiGAN~\cite{Tan_2020}. {\revii Examples of the images generated using these methods are shown in Fig.~\ref{fig:comparison}.} Users were presented with each image in a random order
	(ours on the left and the other method on the right,  or with ours on the right
	and the other method on the left). The reference images were also shown at 10\%
	the size of the synthesized images. The user interface allowed participants to
	zoom in and inspect details of the image, and our instructions encouraged them
	to do so. Each user was asked to indicate which image combined the face of one
	image and the hair of another with the highest quality, and fewest artifacts. On
	average, users spent 90 seconds comparing images before making a selection. We
	asked 396 participants to compare ours to LOHO, and our approach was selected
	378 times (95\%) and LOHO was selected 18 times (5\%).   We asked another 396
	participants to compare against MichiGAN, and the results were 381 (96\%) ours
	vs 14 (4\%) MichiGAN.   The results in both case are statistically significant.

	\subsubsection{Reconstruction Quality}
	\label{sec:reconstruction}
	
	In this work, we measure the reconstruction quality of an embedding using
	various established metrics: RMSE, PSNR, SSIM, VGG perceptual
	similarity~\cite{simonyan2014very}, LPIPS perceptual similarity, and the
	FID~\cite{heusel2017gans} score between the input and embedded images. The
	results are shown in Table~\ref{tab:reconstruction-comparison}.
	
	\begin{table}
	    \caption{A comparison of our method to different algorithms using
			established metrics. Our method achieves the best scores in all metrics.}
		\centering
		%\begin{tabular}{c|cccccc}
		
		\begin{tabularx}{\columnwidth}{X|c@{\hspace{6pt}}c@{\hspace{6pt}}c@{\hspace{6pt}}c@{\hspace{6pt}}c@{\hspace{6pt}}c}
			& RMSE$\downarrow$ & PSNR$\uparrow$ & SSIM$\uparrow$  & VGG$\downarrow$
			& LPIPS$\downarrow$ & FID$\downarrow$ \\
			Baseline & 0.07 & 23.53 & 0.83 & 0.76 & 0.20 & 43.99 \\
			LOHO  & 0.10 & 22.28 & 0.83 & 0.71 & 0.18 & 56.31 \\
			MichiGAN  & 0.06 & 26.51 & 0.88 & 0.48 & 0.12 & 26.82 \\
			Ours & \textbf{0.03} & \textbf{29.91} & \textbf{0.90} & \textbf{0.38} &
			\textbf{0.06} & \textbf{21.21} \\
		\end{tabularx}
		\label{tab:reconstruction-comparison}
	\end{table}

	\subsection{Ablation Study}\label{sec:ablation-study}
	
	\begin{figure}[htb]
		\centering
		\includegraphics[width=1.0\columnwidth]{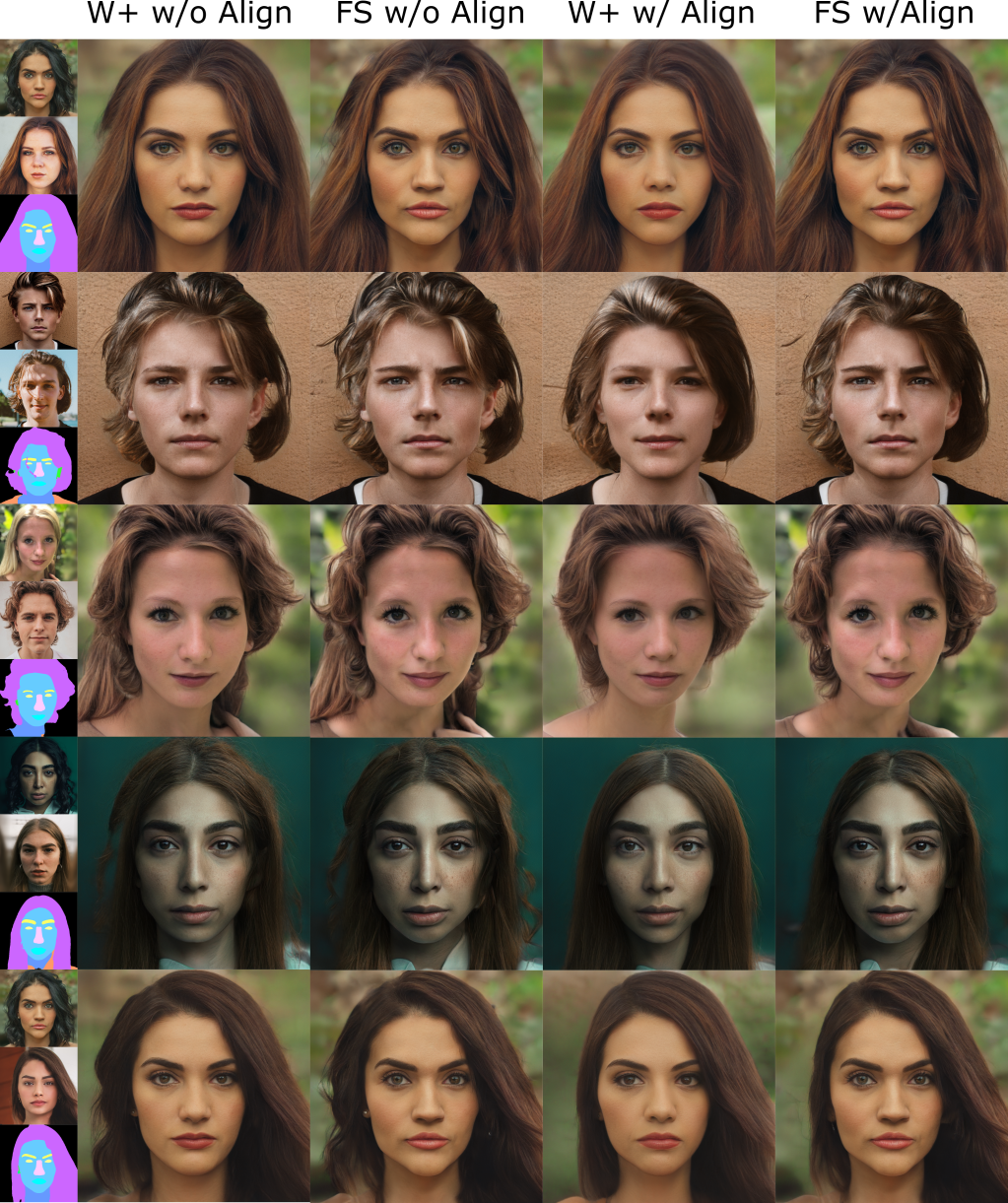}
		\caption{{\revii A qualitative ablation study.
		We compare a baseline version that blends latent codes without image alignment and using $\StructureTensorKInit$, e.g. it uses a $W+$ code (first column), a version that uses fully optimized $FS$ codes, e.g. $\StructureTensorKRec$, but no alignment (second column), $W+$ latent codes with the alignment step (third  column), and our
			complete approach  which uses both alignment and the $FS$ code.  The reference images for the face, hairstyle,
			and the target mask are shown top-to-bottom on the left of each row. This figure shows that alignment ensures that each location has the same semantic meaning (e.g. background), so that inpainting is unnecessary.}}
		\label{fig:qualitative-ablation-study}
	\end{figure}
	
	{\reviii We present a qualitative ablation study of the proposed approach for hairstyle
	transfer. Fig.~\ref{fig:qualitative-ablation-study} provides a visual comparison
	of the results of hairstyle transfer. 
	 A \textit{baseline} version of our
	approach does not include the $FS$ latent space and does not do image alignment
	and is shown in Fig.~\ref{fig:qualitative-ablation-study} and is labeled `$W+$ w/o Align'. It does
	solve for interpolated blending weights to minimize the masked loss function
	from equation~\eqref{eq:l-masked}, however a mixture of unaligned latent codes
	does not always result in a plausible image. This is apparent when {\revii we} compare
	the references image to the synthesized images. Without alignment,  disoccluded regions where the hair region shrinks are not handled properly, and artifacts are visible near the boundary of the hair region.
	The second column uses $\StructureTensorKRec$ rather than $\StructureTensorKAlign$ when blending and so it captures more detail from the original images, however issues caused by lack of semantic alignment remain.   
	The third column of Fig.~\ref{fig:qualitative-ablation-study} includes alignment,
	but it does not use $FS$ space. Without the additional capacity, the
	reconstructed images are biased towards a generic face and hair images, with more symmetry
	and less expression, character, and identifying details than the reference
	images.}
	Overall, the qualitative examples show that each successive modification to the
	proposed approach resulted in higher quality composite images.

	\subsection{Qualitative Results}
	
	In this subsection, we discuss various qualitative results that can be achieved
	using our method.

	{\reviii In Fig.~\ref{fig:hair-gallery} we show many examples of hair style transfer where the structure, shape, and appearance of hair each come from different sources; every row has the same shape and structure, every column has the same appearance. This demonstrates} that our framework can generate a
	large variety of edits.
	Starting from an initial photograph, 
	a user can manipulate a semantic segmentation mask manually to 
	change semantic regions, 
	copy segmented regions from reference images, 
	copy structure information for semantic regions from reference images, 
	and copy appearance information from reference images. 
	In the figure, we show many results where 
	the shape of the hair, 
	the structure of the hair, 
	and the appearance of the hair 
	is copied from three difference reference images.
	Together with the source image, that means that information from up to four
	images contributes to one final blended result image.
	
	In Fig.~\ref{fig:face-swapping} we demonstrate that our framework can handle
	edits to other semantic regions different from the hair. 
	We show how individual facial features such as eyes and eyebrows can be
	transferred from other reference images, how all facial regions can be copied,
	and how all facial regions as well as the appearance can be transferred from
	other source images. We can also attain high quality results for such edits. We
	would like to remark that these edits are generally easier to perform than hair
	transfer.
	
	In Fig.~\ref{fig:comparison} we show selected examples to illustrate why our
	method is strongly preferred compared to the state of the art by users in the
	user study. While previous methods give good results to this very challenging
	problem, we can still achieve significant improvements in multiple aspects.
	First, one can carefully investigate the transition regions between hair and
	either the background and the face to see that previous work often creates hard
	transitions, too similar to copy and pasting regions directly. 
	Our method is able to better make use of the knowledge encoded in GAN latent
	space to find semantic transitions between images.
	Second, other methods can easily create artifacts, due to misalignment in
	reference images. 
	This manifests itself for example in features, e.g. hair structure, being cut
	off unnaturally at the hair boundary. 
	Third, our method achieves a better overall integration of global aspects such
	as lighting.
	The mismatch in lighting also contributes to lower quality transitions between
	hair regions and other regions in other methods. 
	By contrast, other methods also have some advantages over our method. 
	Previous work is better in preserving some background pixels by design. 
	However, this inherently lowers the quality of the transition regions.
	We only focus on hair editing for the comparison, because it seems to be by far
	the most challenging task. This is due to the possible disocclusion of
	background and face regions, the more challenging semantic blending of
	boundaries, and the consistency with global aspects such as lighting. 
	Overall, we believe that we propose a significant improvement to the state of
	the art, as supported by our user study. We also submit all images used in the
	user study as supplementary materials to enable {\revii readers} to inspect the quality
	of our results.
	
	\subsection{Limitations}
\begin{figure}
		\centering
		\includegraphics[width=\linewidth]{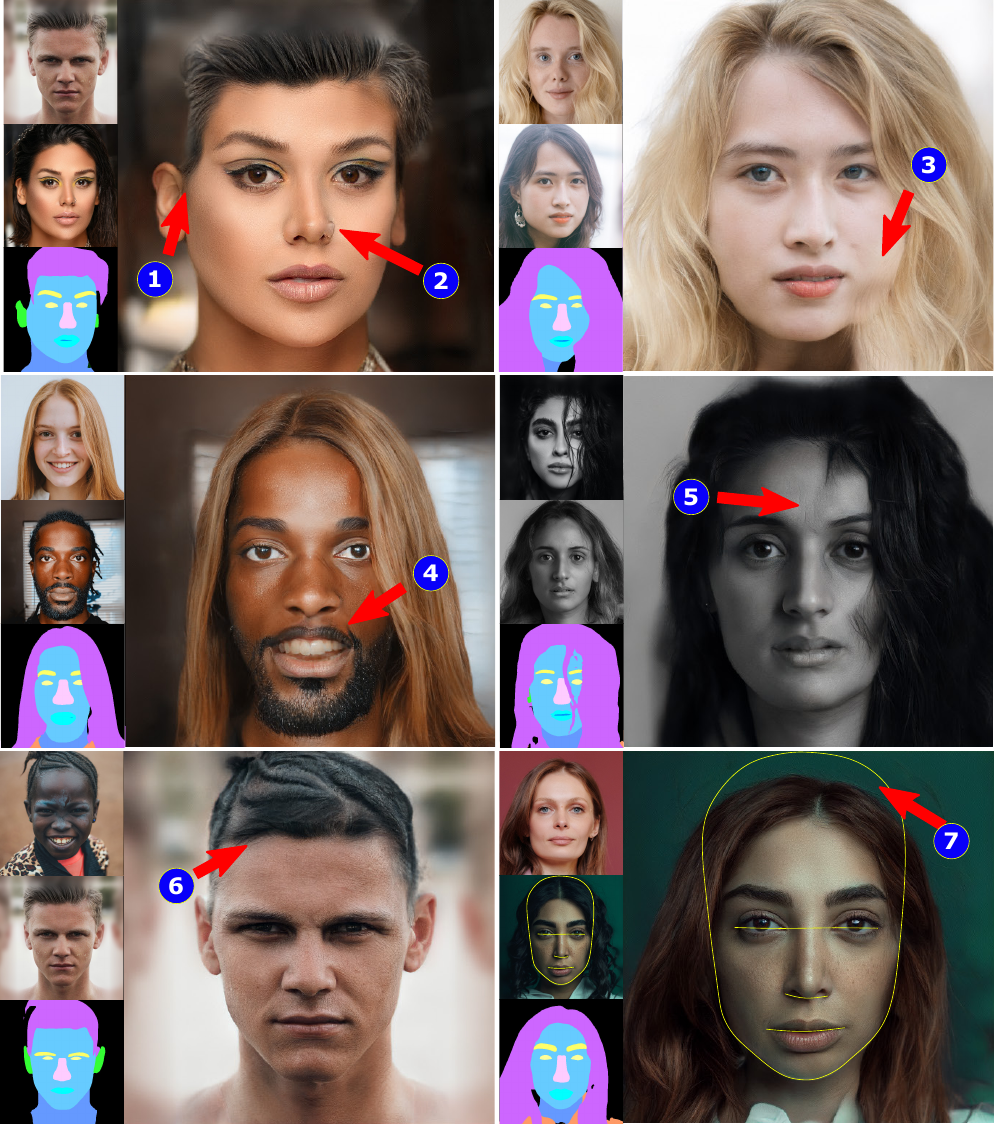}
		\caption{Failure modes of our approach; (1) misaligned segmentation masks
			lead to implausible images; (2, 4) the GAN fails to reconstruct the face,
			replacing lips with teeth or removing jewelry ; (3,5) overlapping translucent or
			thin wisps of hair and face pose a challenge; (6) a region of the target mask
			that is not covered by $\beta_k$ in the hair image is synthesized with a
			different structure; (7) combining images taken from different perspectives can
			produce anatomically unlikely results, the original shape of the head is
			indicated in yellow. }
		\label{fig:limitations}
	\end{figure}

	Our method also has multiple limitations. Even though we increased the capacity
	of the latent space, it is difficult to reconstruct underrepresented features
	from the latent space such as jewelry indicated
	in~Fig.\ref{fig:limitations}(2,4). 
	Second, issues such as occlusion can produce confusing results.  For example,
	thin wisps of hair which also partially reveal the underlying face are difficult
	to capture in Fig.~\ref{fig:limitations}(3,5).
	Many details such as the hair structure in Fig.~\ref{fig:limitations}(7) are
	difficult to preserve when aligning embeddings, and when the reference and
	target segmentation masks do not overlap perfectly the method may fall back to a
	smoother structure. 
	Finally, while our method is tolerant of some errors in the segmentation mask
	input, large geometric distortions cannot be compensated. In
	Fig.~\ref{fig:limitations}(2,7) we show two such examples.
	
	These limitations could be addressed in future work by filtering-out unmatched
	segmentation as was done by LOHO~\cite{saha2021loho}, or by geometrically
	aligning the segmentation masks \textit{before} attempting to transfer the hair
	shape using regularization to keep the segmentation masks plausible and avoid
	issues such as Fig.~\ref{fig:limitations}(1,7). The details of the structure
	tensor could be warped to match the target segmentation to avoid issues such as
	Fig.~\ref{fig:limitations}(6). Issues of thin or transparent occlusions are more
	challenging and may require more capacity or less regularization when finding
	embeddings. 
	
	\section{Conclusions}
	We introduced Barbershop, a novel framework for GAN-based image editing. A user
	of our framework can interact with images by manipulating segmentation masks and
	copying content from different reference images.
	We presented several important novel components. First, we proposed a new latent
	space that combines the commonly used $W+$ style code with a structure tensor.
	% There are so many codes it gets confusing what we mean - W+ is clearer to me. 
	The use of the structure tensor makes the latent code more spatially aware and
	enables us to preserve more facial details during editing.
	Second, we proposed a new GAN-embedding algorithm for aligned embedding. Similar
	to previous work, the algorithm can embed an image to be similar to an input
	image. In addition, the image can be slightly modified to conform to a new
	segmentation mask.
	Third, we propose a novel image compositing algorithm that can blend multiple
	images encoded in our new latent space to yield a high quality result.
	Our results show significant improvements over the current state of the art. In
	a user study, our results are preferred over 95 percent of the time.

	% DO NOT INCLUDE ACKNOWLEDGMENTS IN AN ANONYMOUS SUBMISSION TO SIGGRAPH 2019
	\begin{acks}
	
	We would also like to thank the anonymous reviewers for their insightful comments and constructive remarks.
	This work was supported by the KAUST Office of Sponsored Research (OSR) and the KAUST Visual Computing Center (VCC).

	\end{acks}
	
	% Bibliography
	
	\bibliographystyle{ACM-Reference-Format}
	\bibliography{bibliography}
	
\ifarXive
	% Appendix
	\appendix
	\section{Appendices}
\subsection{Inpainting Masks}

The segmentation masks of a pair of reference images will not always completely cover the entire target segmentation
mask. In this case, there are some uncovered regions, indicated in white in Fig.~\ref{fig:inpaint_mask}(a), that need to be
in-painted in order to create a complete target mask. 
In addition, the hair region is complicated in that some portions of the hair belong \textit{behind} the figure, and
some portions of hair should occlude the figure.  An example of our approach both \textit{without} and \textit{with}
inpainting is shown in Fig.~\ref{fig:Supp_translation} - note that without inpainting we labeled uncovered pixels as
\textit{background} which could cause the background to show through the hair where it should not (middle row). It is
interesting to us that the alignment process, which uses the StyleGAN W+ space as a prior, does not seem to have the
capacity to match the erroneous background regions in the target masks near the subject's forehead. 

When dealing with hair transfer, it is useful to relabel the segmentation masks using three labels of \textit{hair},
\textit{background}, and \textit{other}, where the last label includes the skin, eyes, nose, clothing, etc. We create
the following masks: $\Mask_\text{hair}^\text{behind}$ is a mask labeled as \textit{background} wherever both
references were background, and labeled as \textit{hair} wherever the reference image for the hair was labeled as
\textit{hair}.  The remaining pixels of $\Mask_\text{hair}^\text{behind}$ are unknown, and they may be portions of
hair that pass behind the subject. Therefore, we inpaint $\Mask_\text{hair}^\text{behind}$ using the fast-marching
method of \cite{telea2004image}, which is implemented by OpenCV. Next, we create a mask
$\Mask_\text{other}^\text{middle}$ using the segmentation regions of the \textit{other} reference image, except
that its original \textit{hair} region is inpainted using the same exact approach.  
Finally, we construct a mask  $\Mask$ in three layers: we first initialize the mask with
$\Mask_\text{hair}^\text{behind},$  and then we transfer the labels \textit{other} than background from
$\Mask_\text{other}^\text{middle}$, and finally we set any pixel that was hair in the original reference image for
hair to also have the label \textit{hair}  in $\Mask$ so that we retain the bangs, or locks of hair which pass in
front of the face or shoulders.  The heuristic approach we used is not capable of generating segmentation masks for
completely occluded features (such as eyes or ears) that were covered by hair, however GAN-based inpainting approaches
for the masks themselves are a subject of future work.

\subsection{Sensitivity to Pose and Spatial Alignment}

The proposed approach works for cropped portrait images -- these images are always somewhat aligned, with a single
dominant face in the center and a frontal or three-quarters vies of a face. This is both due to a preference for this by
photographers, but also that the datasets are collected by automatically cropping the images using a facial alignment
net such as DLIB or FAN~\cite{bulat2017far}.  The use of face and pose detection networks could allow one to filter out
incompatible reference-images and thus mitigate issues with spatial alignment. We did not do this filtering in our user
study, so errors caused by misalignment were included in our evaluations.  It is therefore useful to understand how
sensitive the proposed approach is to  changes in the spatial alignments of reference images.  In order to demonstrate
the qualitative effect of our approach to misalignment of the mask, we translated the hair region in
Fig.~\ref{fig:Supp_translation} when generating the target mask. 

\subsection{Manually Editing Masks}

The main focus of this paper is on completely automated hair transfer, however it is possible to overcome many of the
challenges and limitations of an automatic approach if user edits are allowed.  For example, by allowing a very limited
set of user interactions (dragging, scaling, and flipping) of the hair region we can achieve results shown in
Fig.~\ref{fig:Supp_manual}.

\subsection{Comparison to Concurrent Work}
Concurrently with our work, StyleMapGan~\cite{kim2021stylemapgan} is also capable of doing face transfer. We compare
against prior work on this task in 
Fig.~7 %\ref{fig:face-swapping}. 
We illustrate the differences between our results and
StyleMapGan qualitatively in Fig.~\ref{fig:compare_with_StylemapGAN}, which demonstrates the results of eyes and
eyebrows (top row) and entire faced (bottom row). We observe that our proposed solution is capable of preserving the
details of the composited parts.

\begin{figure}[thpb]
	\centering
	\includegraphics[width=\linewidth]{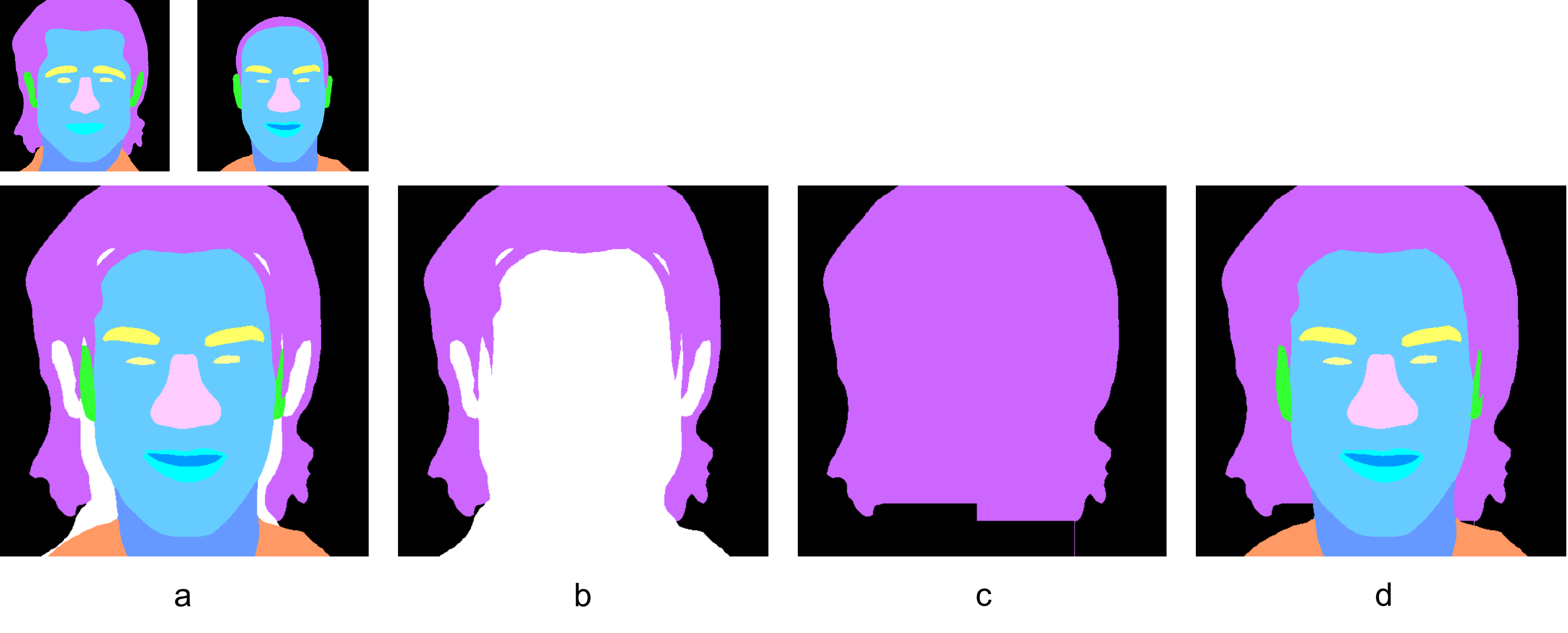}
	\caption{Mask inpainting. {\revii The semantic segmentations of two reference images for \textit{hair} and \textit{other} are shown on the first row. The second row shows (a) a composite mask without inpainting and disoccluded pixels shown in white; (b) the hair region before inpainting, (c) the result of inpainting the hair mask; (d) the result of filling-in disoccluded regions of (a) using the mask from (c). } }
	\label{fig:inpaint_mask}
\end{figure}

\begin{figure*}[thpb]
	\centering
	\includegraphics[width=\linewidth]{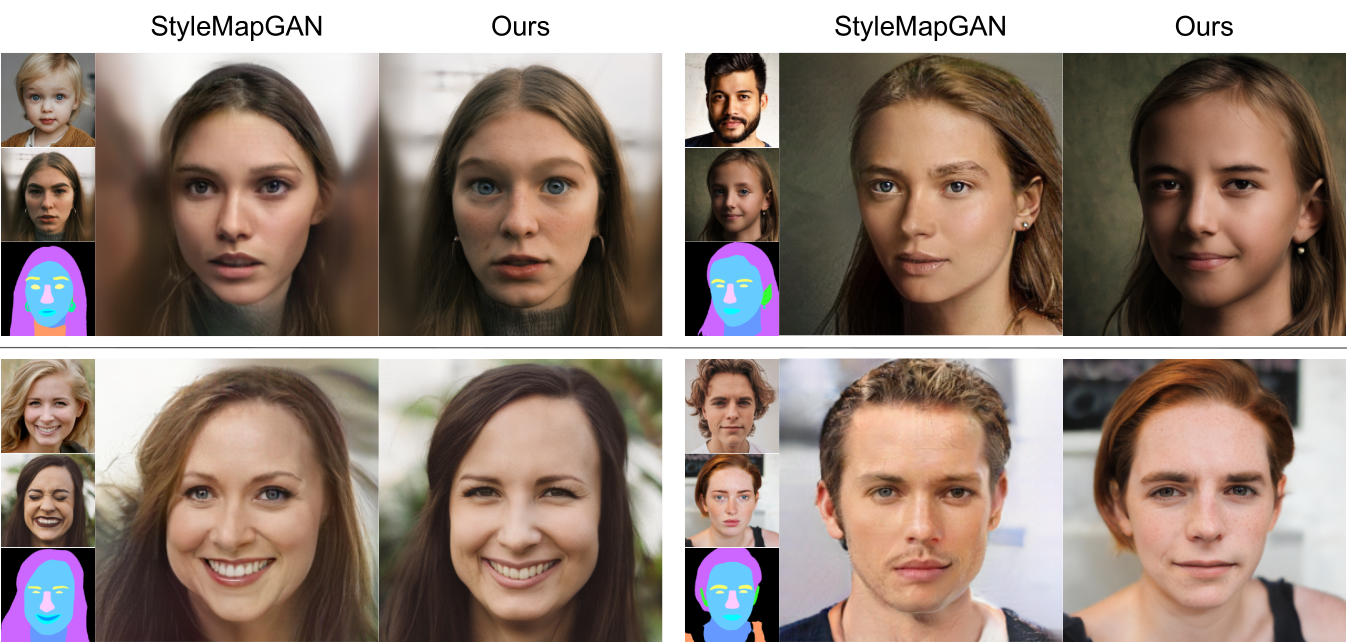}
	\caption{Comparison with StyleMapGAN~\cite{kim2021stylemapgan}. First row: examples of eye and eyebrow transfer;
		second row: examples of face swapping. Note that ours successfully edits the portrait locally, while StyleMapGAN
		provides a completely different person.}
	\label{fig:compare_with_StylemapGAN}
\end{figure*}

\begin{figure*}[thpb]
	\centering
	\includegraphics[width=\linewidth]{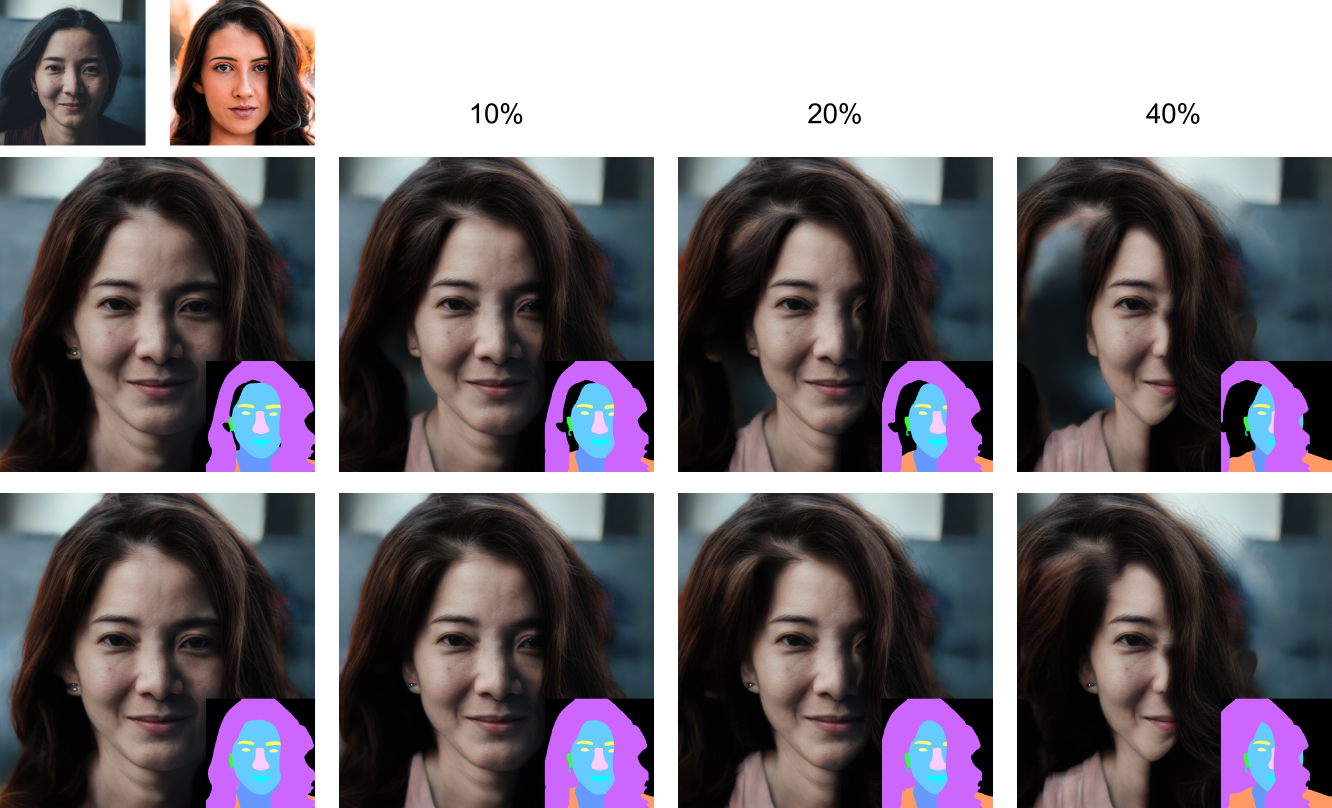}
	\caption{Misalgin the target segmentation mask by shifting. First row: translate the target hair region without
		preprocessing the segmentation mask; Second row: use the segmentation mask preprocessing step. Please note the artifacts
		between the hair and neck.}
	\label{fig:Supp_translation}
\end{figure*}

\begin{figure*}[thpb]
	\centering
	\includegraphics[width=\linewidth]{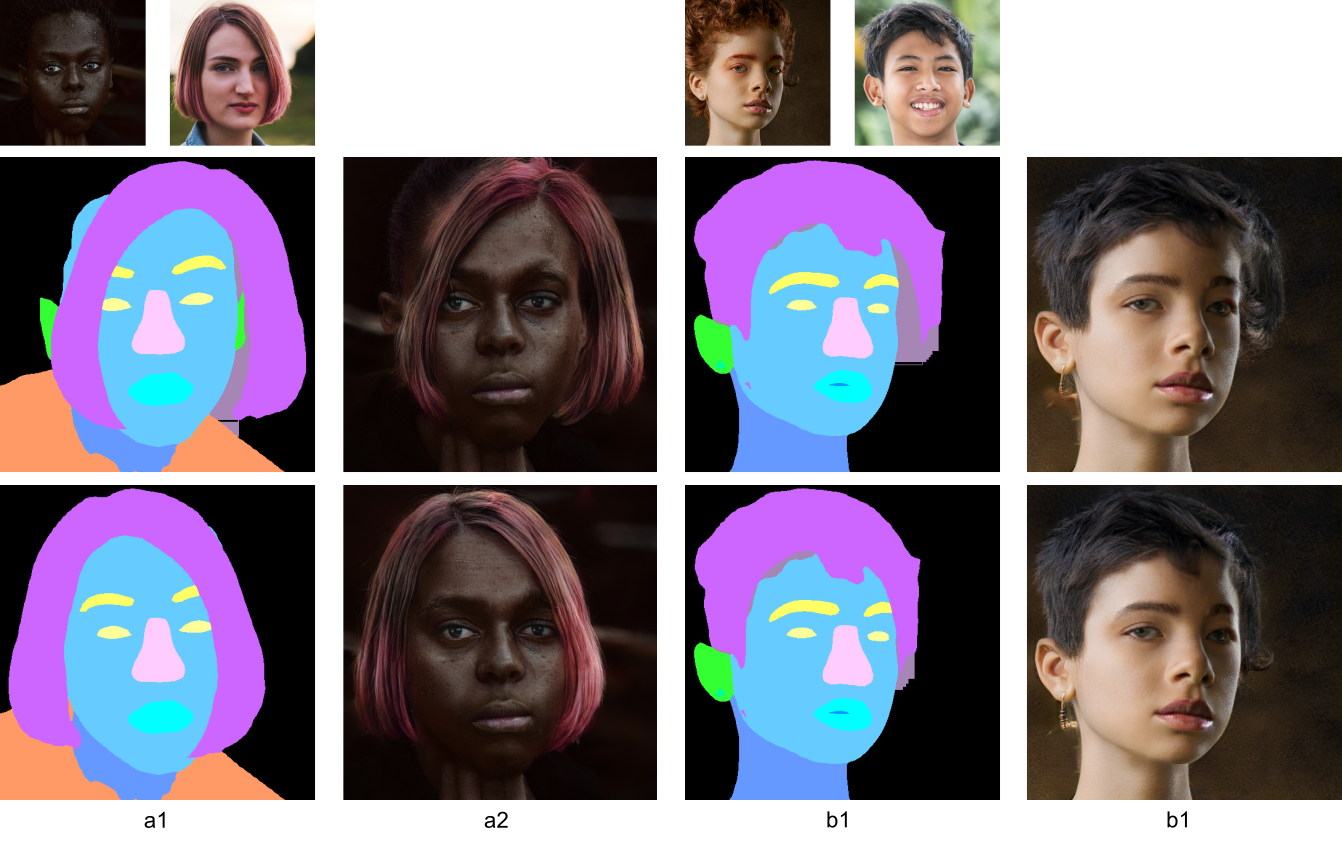}
	\caption{The second row shows the results of manually editing the target segmentation mask. The left portion of the
		figure shows an example where the hair and the face could be aligned by flipping the hair segmentation mask. The right
		portion shows an example in which the regions could be better aligned by translating them}
	\label{fig:Supp_manual}
\end{figure*}

\fi

	% \section{Switching Times}
	
	% In this appendix, we measure the channel switching time of Micaz
	% \cite{CROSSBOW} sensor devices.  In our experiments, one mote
	% alternatingly switches between Channels~11 and~12. Every time after
	% the node switches to a channel, it sends out a packet immediately and
	% then changes to a new channel as soon as the transmission is finished.
	% We measure the number of packets the test mote can send in 10 seconds,
	% denoted as $N_{1}$. In contrast, we also measure the same value of the
	% test mote without switching channels, denoted as $N_{2}$. We calculate
	% the channel-switching time $s$ as
	% \begin{displaymath}%
	% s=\frac{10}{N_{1}}-\frac{10}{N_{2}}.
	% \end{displaymath}%
	% By repeating the experiments 100 times, we get the average
	% channel-switching time of Micaz motes: 24.3\,$\mu$s.

\end{document}